\documentclass[preprint,12pt,authoryear]{elsarticle}

\usepackage{lineno}
\usepackage{amssymb}

\usepackage{amsfonts}
\usepackage{amsmath}
\usepackage{booktabs}
\usepackage{multirow}
\usepackage{pbox}
\usepackage{tikz}
\usepackage{booktabs}
\usepackage{rotating}
\usepackage[]{graphicx}
\usepackage{fontawesome5}
\usepackage{adjustbox}
\usepackage{orcidlink}

\definecolor{gold}{rgb}{1.0, 0.84, 0.0}
\definecolor{silver}{rgb}{0.75, 0.75, 0.75}
\definecolor{bronze}{rgb}{0.8, 0.5, 0.2}

\newcommand{\goldtrophy}{\textcolor{gold}{\faTrophy}}
\newcommand{\silvertrophy}{\textcolor{silver}{\faTrophy}}
\newcommand{\bronzetrophy}{\textcolor{bronze}{\faTrophy}}
\newcommand{\fourthtrophy}{\textcolor{black!75}{\faTrophy}}

\tikzset{every node/.append style={text depth=0.4ex}}
\usetikzlibrary{calc}

\usepackage{hyperref}
\usepackage{bookmark}
\usepackage{caption}
\usepackage{apalike}

\begin{document}

\begin{frontmatter}

\title{Few-shot multi-token DreamBooth with LoRa for style-consistent character generation} 

\author[label]{Ruben~Pascual \orcidlink{0009-0002-5250-8554}}\ead{ruben.pascual@unavarra.es}
\author[label]{Mikel~Sesma-Sara \orcidlink{0000-0002-2949-7909}}\ead{mikel.sesma@unavarra.es}
\author[label]{Aranzazu~Jurio \orcidlink{0000-0002-4087-586X}}\ead{aranzazu.jurio@unavarra.es}
\author[label]{Daniel~Paternain \orcidlink{0000-0002-5845-887X}}\ead{daniel.paternain@unavarra.es}
\author[label]{Mikel~Galar \orcidlink{0000-0003-2865-6549}}\ead{mikel.galar@unavarra.es}

\affiliation[label]{organization={Institute of Smart Cities (ISC) and Department of Statistics, Computer Science and Mathematics, Public University of Navarre (UPNA)},
            addressline={Campus Arrosadia}, 
            city={Pamplona},
            postcode={31006}, 
            state={Navarre},
            country={Spain}}

\begin{abstract}

The audiovisual industry is undergoing a profound transformation as it is integrating AI developments not only to automate routine tasks but also to inspire new forms of art. This paper addresses the problem of producing a virtually unlimited number of novel characters that preserve the artistic style and shared visual traits of a small set of human-designed reference characters, thus broadening creative possibilities in animation, gaming, and related domains. Our solution builds upon DreamBooth, a well-established fine-tuning technique for text-to-image diffusion models, and adapts it to tackle two core challenges: capturing intricate character details beyond textual prompts and the few-shot nature of the training data. To achieve this, we propose a multi-token strategy, using clustering to assign separate tokens to individual characters and their collective style, combined with LoRA-based parameter-efficient fine-tuning. By removing the class-specific regularization set and introducing random tokens and embeddings during generation, our approach allows for unlimited character creation while preserving the learned style. We evaluate our method on five small specialized datasets, comparing it to relevant baselines using both quantitative metrics and a human evaluation study.  Our results demonstrate that our approach produces high-quality, diverse characters while preserving the distinctive aesthetic features of the reference characters, with human evaluation further reinforcing its effectiveness and highlighting the potential of our method.

\end{abstract}

\begin{keyword}
Text-to-image diffusion models \sep DreamBooth \sep Character generation \sep Few-shot
\end{keyword}

\end{frontmatter}

\begin{figure*}[!ht]
    \centering
    \includegraphics[width=0.75\linewidth]{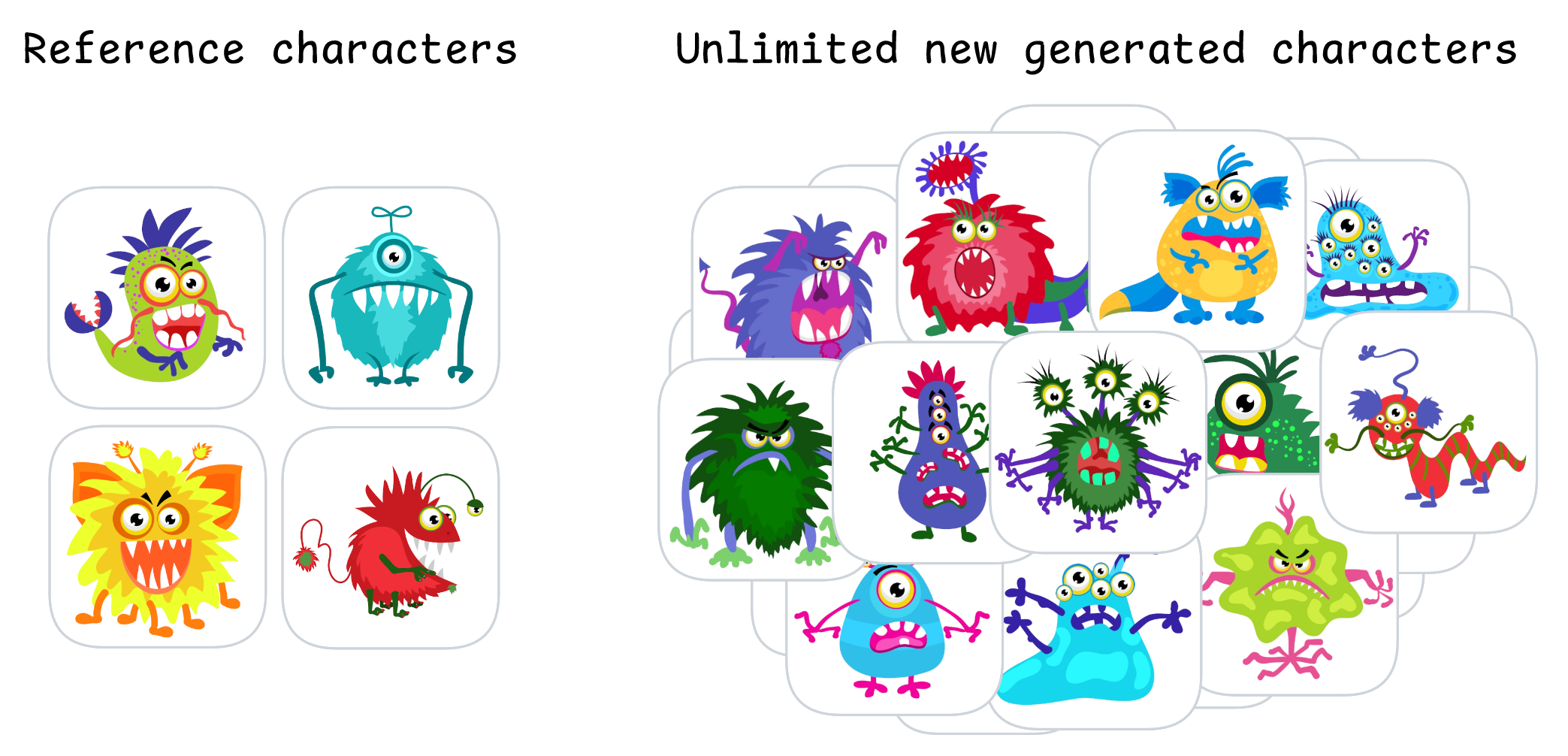}
    \caption{Visual example illustrating the problem (images taken from scary dataset \citep{macrovector2023ScaryD}). Starting from a limited set of human-designed reference characters (left), the goal is to generate new characters that maintain the same style and visual characteristics (right).}
    \label{fig:problemDef}
\end{figure*}

\section{Introduction}

Artificial Intelligence is reshaping the audiovisual industry, affecting both creative and technical workflows. In video games, AI-powered tools like NVIDIA DLSS~\citep{lin2025NVIDIAD4IMFGEADT} improve real-time rendering. In film and television, AI is transforming multiple stages of production. Large Language Models (LLMs), such as Dramatron~\citep{mirowski2023CoWritingSTSLMEIP}, assist in scriptwriting by generating structured narratives, while post-production benefits from tools such as Adobe's Generative Fill~\citep{clark2024NextGGANPAB}, which simplifies complex edits. The integration of AI in CGI has also enabled de-aging effects~\citep{edwards202450MMHDTHGA} and even digital resurrection of actors~\citep{sherwood2025DigitalRFFRD, thurston2024IanHWBDNHAR}. 

As these AI-driven advancements transform filmmaking and interactive media, one particularly complex challenge remains: character generation. Creating unique yet stylistically consistent characters is essential in animation, gaming, and visual storytelling. Traditionally, artists manually designed an array of characters that fit within a cohesive artistic style, a process that demands both creativity and time. Over time, various attempts have been made to automate this process. For instance, video games like \textit{The Sims} offer character creation systems, allowing users to customize avatars by selecting predefined features such as facial structures, hairstyles and attire~\citep{leigh2014Sims4PG}. Additionally, procedural generation techniques have been employed to algorithmically create diverse characters~\citep{khatchadourian2015GalaxySizedVG}. More recently, advancements in AI have led to the development of models capable of generating characters. For example, Generative Adversarial Networks (GANs) have been utilized to create anime characters~\citep{ruan2022AnimeCGGAN} and platforms like Artguru offer AI-powered tools to generate custom Pokémon-like creatures based on textual descriptions~\citep{artguru2025AIPG}. However, both approaches benefit from extensive datasets of well-known styles like anime and Pokémon, which helps them to effectively learn visual traits. Developing models capable of generating stylistically consistent characters from limited reference images remains an ongoing challenge.

This work addresses this specific issue. As illustrated in Figure \ref{fig:problemDef}, the task is to generate an arbitrary number of new characters (right) that preserve the style and visual consistency of a set of human-designed reference characters (left). However, this is a complex problem due to several significant challenges. Firstly, the target style is not predefined or widely recognized, nor is it necessarily easily describable with a textual prompt. Instead, it must be learned directly from a potentially limited set of reference images, making the task inherently few-shot~\citep{snell2017PrototypicalNFL}. Additionally, beyond maintaining a consistent style, the generated characters must incorporate novel elements to ensure that they are not mere replicas of the references.

Furthermore, objectively evaluating the results in this scenario is particularly challenging, as the limited size of the reference data and the inherent subjectivity of the task make it difficult for standard metrics~\citep{hessel2021CLIPScoreREMIC, heusel2017GANsTTTURCLNE,sajjadi2018AssessingGMPR,salimans2016ImprovedTTG,wang2004ImageQAEVSS,zhang2018UnreasonableEDFPM} to capture subtle stylistic differences, which complicates reliable assessment.

Interestingly, this challenge is not only tied to character generation, but also relates to broader machine learning problems, particularly class imbalance~\citep{fernandez2018LearningIDS}. When data for certain categories are scarce, models often struggle to generalize, leading to biased predictions. In such cases, being able to generalize from few-shot examples in image generation can serve as a form of data augmentation, helping to mitigate imbalances.

To address this challenge, there is a wide variety of image generation models, ranging from open-source to proprietary solutions, with some of the most renowned being DALL-E 3~\citep{betker2023ImprovingIGBC}, Midjourney~\citep{midjourney2024}, Flux~\citep{2025FLUX} and Stable Diffusion~\citep{rombach2022HighResolutionISLDM}, including its various versions~\citep{moser2024DiffusionMISES,podell2023SDXLILDMHIS}. However, achieving precise control over the output of these models can be very challenging, but it is essential for the task at hand to ensure that generated characters maintain both stylistic consistency and sufficient diversity. 

Several methodologies have been proposed to gain control over generative models~\citep{avrahami2023SpaTextSRCIG,gal2022ImageWOWPTGUTI,google2025NanoBanana,huang2025DiffStylerCDDTIS,sohn2023StyleDropTGAS,zhang2023AddingCCTDM}. Some approaches enhance diffusion models through structured visual conditioning, such as edge maps and spatial constraints \citep{avrahami2023SpaTextSRCIG,zhang2023AddingCCTDM}, while other refine text-to-image generation by embedding new concepts \citep{gal2022ImageWOWPTGUTI}, adapting images to match specific artistic styles described by text \citep{huang2025DiffStylerCDDTIS}, or adapting models to mimic artistic styles directly from reference images \citep{sohn2023StyleDropTGAS}. Although effective in certain contexts, these methods often require explicit textual descriptions, iterative refinements, or external conditioning, making them less suitable for generating unlimited stylistically consistent characters from a few reference images. Among these methods, DreamBooth~\citep{ruiz2023DreamBoothFTTDMSG} fine-tunes a pre-trained model by associating a new subject with a previously unused token, enabling the integration of new concepts. However, its reliance on textual prompts constrains its ability to generate a diverse set of characters. Despite this limitation, its ability to integrate new visual concepts positions it as a natural foundation for our approach.

For our proposal, we build upon DreamBooth, adapting it to improve both the learning of stylistic and character features, as well as the generation diversity. However, as shown in our previous work~\citep{pascual2024EnhancingDLGUCSD}, directly applying DreamBooth to our task presents some challenges: it involves high computational costs and does not fully preserve stylistic consistency across generations. To address these issues, we introduced LoRa-based fine-tuning to reduce computational overhead, a multi-token strategy to better distinguish individual character identity from their shared artistic style, and the removal of the regularization dataset to encourage style adaptation. Additionally, we incorporated random tokens and embeddings during generation to increase character diversity without compromising stylistic coherence.
 
In this work, we extend our previous approach by refining the methodology and substantially expanding the experimental evaluation. Methodologically, we introduce clustering to improve the multi-token strategy, allowing better separation of character-specific features from global stylistic traits, and we adopt a random multivariate embedding sampling technique for generation, improving diversity and coherence. Experimentally, we evaluate our method on five specialized datasets using quantitative metrics, including an adapted style preservation measure (Fidelity), and a large-scale human study comprising 3,920 pairwise comparisons from 135 general participants and 300 expert ratings from six professional artists.

In summary, the key contributions of this work are:
\begin{itemize}
    \item A clustering-based multi-token strategy for separating identity and style features.
    \item A multivariate embedding sampling method for diverse and coherent generation. 
    \item A style-adapted metric (Fidelity) for evaluating style preservation.
    \item A large-scale human evaluation combining general and expert feedback.
\end{itemize}

These contributions offer a sound framework for generating unlimited, style-consistent characters from limited visual references, with results validated through both quantitative metrics and human evaluation.

\section{Background}\label{sec-background}

This section reviews related work on image generation, fine-tuning techniques, and evaluation methods, highlighting their strengths and limitations for our task (Section \ref{subsec-related}), and provides a detailed discussion on DreamBooth, the fine-tuning method that serves as the foundation of our approach (Section \ref{subsec-dreamboothMethod}).

\subsection{Related work}\label{subsec-related}

\vspace{6pt}\noindent \textbf{Image Generation Methods}. 
Generative models have evolved significantly over the past decade. Variational Autoencoders (VAEs) \citep{kingma2013AutoEncodingVB} introduced probabilistic latent representations but lacked fine-grained detail \citep{razavi2019GeneratingDHIV}. Generative Adversarial Networks (GANs) \citep{goodfellow2020GenerativeAN} improved quality and achieved success across various tasks \citep{karras2019StyleBasedGAGAN, ledig2017PhotoRealisticSISUGAN, wang2020ImaGINatorCSGVG}, yet remained highly dependent on large training datasets, limiting their applicability in data-scarce scenarios \citep{karras2020TrainingGANLD}.
Recently, research has increasingly turned towards diffusion models \citep{ho2020DenoisingDPM}, which have demonstrated superior generative quality in various generative tasks \citep{dhariwal2021DiffusionMBGIS}. Large-scale text-to-image diffusion models, such as DALL-E 3~\citep{betker2023ImprovingIGBC}, Imagen~\citep{saharia2022PhotorealisticTDMDLU} and Stable Diffusion~\citep{rombach2022HighResolutionISLDM}, have achieved remarkable performance in generating diverse and complex visual content. However, fine-tuning these large pre-trained models to introduce novel concepts or adapt them to specialized tasks remains challenging. The size of their parameter space and the complexity of their pre-trained representations make the adaptation process highly resource-intensive, driving the need for model customization techniques that enable fine-tuned control and concept integration. 

\vspace{6pt}\noindent\textbf{Model customization} techniques have emerged to adapt pre-trained diffusion models for precise visual control and novel concept integration. Explicit visual conditioning techniques such as ControlNet~\citep{zhang2023AddingCCTDM}, SpaText~\citep{avrahami2023SpaTextSRCIG}, LayoutDiffusion~\citep{zheng2023LayoutDiffusionCDMLG}, T2I-Adapter~\citep{mou2024T2IAdapterLADOMCATDM}, and Prompt-to-Prompt\citep{hertz2022PrompttoPromptIECC} condition diffusion models through structural inputs, sketches, or detailed prompt manipulations, offering precise spatial control. More implicit approaches, such as FABRIC~\citep{vonrutte2023FABRICPDMIF} and Diffusion Self-Guidance~\citep{epstein2023DiffusionSCIG}, refine generation aiming at more controlled outputs by iteratively guiding latent representations. 

There are also techniques focused on embedding personalized concepts into diffusion models. Methods like Textual Inversion~\citep{gal2022ImageWOWPTGUTI} and CustomDiffusion~\citep{kumari2023MultiConceptCTD} facilitate targeted personalization by learning novel concepts through specialized textual embeddings. In contrast, StyleDrop~\citep{sohn2023StyleDropTGAS} customizes diffusion models specifically to reproduce artistic styles, but also depends on explicit textual descriptions. Additionally, DreamBooth~\citep{ruiz2023DreamBoothFTTDMSG} fine-tunes diffusion models by associating new visual concepts with unique textual identifiers. Nevertheless, DreamBooth relies on precise textual prompt and class-specific regularization, limiting its suitability for generating diverse and stylistically consistent characters from limited reference data. Moreover, DreamBooth requires full model fine-tuning, resulting in high computational costs, although this limitation can be mitigated using parameter-efficient fine-tuning (PEFT) techniques.

\vspace{6pt}\noindent\textbf{Parameter-Efficient Fine-Tuning (PEFT)}~\citep{xu2023ParameterEfficientFMPLMCRA} methods have been developed to mitigate the computational overhead associated with full model fine-tuning. By reducing memory and storage demands, PEFT strategies enable efficient adaptation of large models while maintaining comparable performance levels. One of the most prominent PEFT techniques is LoRA (Low-Rank Adaptation)~\citep{hu2021LoRALALLM}, which reduces the number of trainable parameters by decomposing the model's weight matrices into smaller, low-rank representations. This significantly reduces the computational burden while preserving the model's ability to learn new concepts. LoRA integrates seamlessly with DreamBooth and has inspired various refinements, including LoHA~\citep{hyeon-woo2021FedParaLHPCFL} and DyLoRA~\citep{valipour2023DyLoRAPTPMUDSLA}. Beyond LoRA, other PEFT methods include prompt tuning~\citep{lester2021PowerSPPT}, prefix tuning~\citep{li2021PrefixTuningOCPG}, and adapter-based approaches~\citep{liu2024GPTUT}. Nevertheless, despite the variety of techniques available, LoRA remains the most widely adopted for fine-tuning models like Stable Diffusion due to its simplicity, computational efficiency and high performance.

\vspace{6pt}\noindent\textbf{Evaluation of generated images.} Ensuring that a generated character maintains stylistic coherence with respect to a given dataset is particularly challenging, as it requires balancing visual quality, diversity, and adherence to the dataset’s artistic characteristics, none of which are fully captured by a single metric. In general, evaluating generative image models is inherently complex due to the subjective nature of visual quality and diversity. Some metrics aim at capturing the quantitative differences between generated and real image distributions. Among these, the Fréchet Inception Distance (FID)~\citep{heusel2017GANsTTTURCLNE} is a widely used standard due to its sensitivity to image realism, although it measures distributional similarity rather than image-level quality. Complementary metrics such as Precision and Recall~\citep{sajjadi2018AssessingGMPR} provide deeper insights by independently quantifying image fidelity (precision) and diversity (recall). Both metrics rely on sufficiently large sample sizes to provide reliable statistical estimates, limiting their applicability in few-shot scenarios. Another frequently used metric is the Inception Score (IS)~\citep{salimans2016ImprovedTTG}, which evaluates generative quality based on class predictions derived from ImageNet categories, but it is unsuitable for our scenario, as it assumes the presence of meaningful semantic classes.

Image-level metrics, comparing individual generated images against references, offer more granular evaluation. The Learned Perceptual Image Patch Similarity (LPIPS)~\citep{zhang2018UnreasonableEDFPM} measures perceptual similarity by directly comparing neural network embeddings. Alternatively, the Structural Similarity Index (SSIM)~\citep{wang2004ImageQAEVSS} is commonly employed to assess structural alignment between paired images, yet is primarily sensitive to low-level distortions rather than high-level stylistic differences. More recently, CLIPScore~\citep{hessel2021CLIPScoreREMIC} and its image–image variant CLIP-I~\citep{ruiz2023DreamBoothFTTDMSG} have gained popularity for evaluating semantic coherence. Both rely on CLIP’s~\citep{radford2021LearningTVMNLS} vision–language embeddings: CLIPScore measures alignment between images and textual prompts, while CLIP-I compares images. However, these metrics are limited for our purposes, as they cannot specifically assess stylistic consistency independently of character identity, which is essential for generating an unlimited variety of unique characters.

Despite these advances, quantitative metrics alone remain insufficient to fully capture subjective dimensions like stylistic coherence, and novelty~\citep{jayasumana2024RethinkingFBEMIG}. Consequently, human evaluations remain indispensable~\citep{otani2023VerifiableRHETG}, typically involving perceptual similarity assessments, rating tasks, or comparative judgments. 

\subsection{DreamBooth method} \label{subsec-dreamboothMethod}

DreamBooth's training process is designed to teach text-to-image diffusion models a new subject (e.g., one's pet dog) and unfolds in two distinct stages: training and generation.

\begin{figure*}[ht!]
    \centering
    \includegraphics[width=\linewidth]{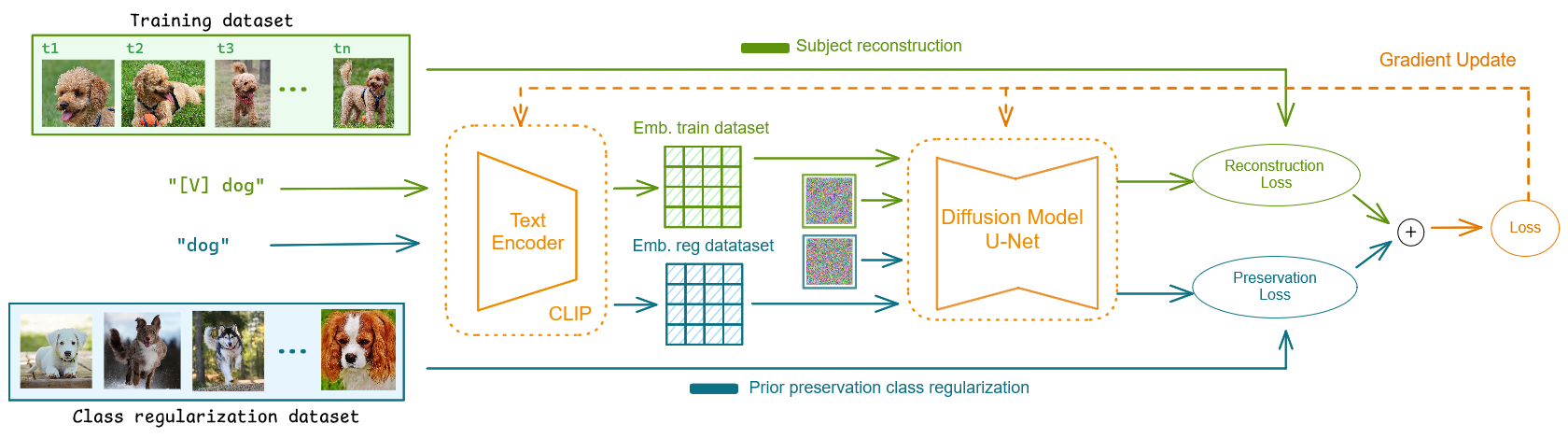}
    \caption{Diagram illustrating Dreambooth training, showcasing an example of training the model to generate images of a specific pet dog.}
    \label{fig:dreambooth}
\end{figure*}

\vspace{6pt}\noindent\textbf{Training.} The training phase of DreamBooth includes two core components:
\begin{enumerate}
    \item \textit{Inserting a novel subject into the model's knowledge.} This component, illustrated at the top of Figure \ref{fig:dreambooth}, focuses on inserting a novel subject into the model's knowledge. This involves using a specific prompt that accompanies each training image in the format 
    \begin{center}``a \texttt{[id] [class]}''\end{center}  
    where
    \begin{itemize}
        \item \texttt{[id]} represents a unique, rarely used token that corresponds exclusively to the subject being learned, and
        \item \texttt{[class]} denotes a general category to which the subject belongs (e.g., ``dog'' if the subject is one's pet dog).
    \end{itemize} 
    The purpose of this structure is to take advantage of the model's pre-existing understanding of \texttt{[class]} features, facilitating the association between the unique identifier \texttt{[id]} and the specific visual features of the new subject. However, this process carries the risk of overwriting the model's prior knowledge of \texttt{[class]} with the specific features of the new subject (e.g., causing the model to only generate images of the specific pet dog whenever prompted with ``dog''). To mitigate this, a regularization procedure is introduced in the second component. 
    \item \textit{Regularization.} This component, depicted at the bottom of Figure \ref{fig:dreambooth}, prevents the model from forgetting its prior knowledge. Prior to training, the model is used to generate a set of regularization images by prompting it simply with the general class descriptor (e.g., ``a dog''), capturing the model's existing understanding of that class. During training, these regularization images are fed into the model alongside the subject-specific images, and their associated loss is combined with that of the training set. 
\end{enumerate}

This dual approach enables DreamBooth to incorporate new subjects while preserving the model’s original comprehension of the broader class, balancing the learning of novel concepts with the preservation of prior knowledge.

\vspace{6pt}\noindent\textbf{Generation.} After training, DreamBooth generation requires only the use of the rare token \texttt{[id]} and the class descriptor \texttt{[class]} in the prompt. The model recognized \texttt{[id]} as the newly trained specific subject and \texttt{[class]} as its general category, enabling controlled image generation. By modifying the prompt with additional details, such as ``A \texttt{[id]} dog on the moon'', the model can produce high-quality images that accurately reflect the target subject in various scenes.

\section{Multi-token DreamBooth with LoRA}\label{sec-proposal}
In this section, we first analyze the key limitations of DreamBooth in this context (Section \ref{subsec-motivation}). We then introduce our proposed method, multi-token DreamBooth with LoRA, which enhances style retention, improves character diversity, and enables scalable character generation (Section \ref{subsec-ourmethod}).

\subsection{Motivation}\label{subsec-motivation}
As a method for personalized image generation, DreamBooth can be used to learn an artistic style rather than a specific subject. In the context of our problem, the rarely used token \texttt{[id]} could be associated with the artistic style and shared visual traits of the characters of the training dataset using the prompt ``\texttt{[id]} style''. Then, during inference, this token would be used to generate new characters that are consistent with the learned style. However, this approach faces two fundamental limitations:

\begin{enumerate}
    \item \textit{Over-specialization to training data}: In this context, where only a few reference images (10-30) are available, DreamBooth is prone to overfitting. Instead of learning a generalized artistic style, the token \texttt{[id]} tends to capture specific character details, resulting in only slight variations of the reference images, rather than novel outputs. Even with LoRa, which has been proven to retain prior knowledge and prevent overfitting~\citep{ojha2021FewshotIGCC}, our experiments show that DreamBooth still struggles to generate diverse characters beyond the training set. The model tends to reproduce familiar patterns.
    \item \textit{Lack of scalability for generating unlimited characters}: DreamBooth does not inherently support scalable character variation as it is designed to learn a single visual concept. Thus, generating multiple unique characters within the same artistic style still requires manual prompt modifications, which contradicts our goal of fully automated unlimited and diverse character generation.
\end{enumerate}

These limitations demonstrate the need for a more effective approach. In the next section, we introduce multi-token DreamBooth with LoRA, designed to enhance style retention, character diversity, and scalability.

\subsection{Our proposal}\label{subsec-ourmethod}

To overcome the aforementioned issues, we propose modifications to the training and generation phases of DreamBooth. 

\begin{figure*}[ht!]

    \centering
    \includegraphics[width=\linewidth]{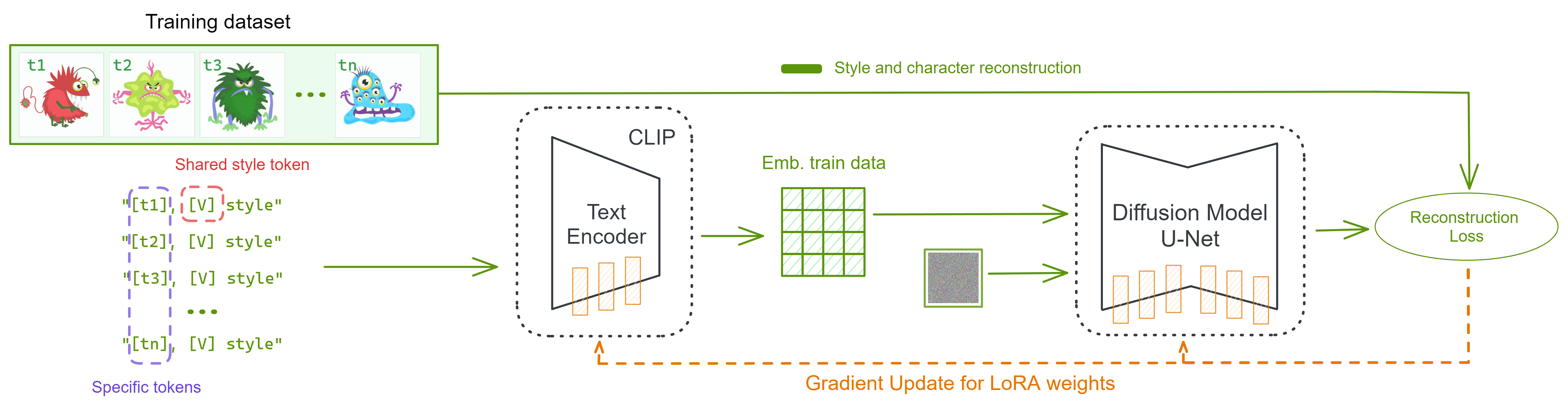}
    \caption{Diagram illustrating proposed multi-token DreamBooth with LoRA training, showcasing an example of training the model for our specific problem.}
    \label{fig:ourmethod}
\end{figure*}

\vspace{6pt}\noindent\textbf{Training.} Our approach introduces three key modifications to DreamBooth training process (see Figure \ref{fig:ourmethod}).

    \vspace{6pt}\noindent1. \textit{Character-specific tokens:} Instead of a single token representing the entire style, we introduce separate tokens for each character alongside a shared token for the overall style. This aims to explicitly disentangle individual character features from the broader artistic style, allowing the model to learn them separately. Thus, training prompts follow the format \begin{center}
        ``\texttt{[specific\_id] [shared\_id] [class]}'',
    \end{center} 
    where:
    \begin{itemize}
        \item \texttt{[specific\_id]}: A unique token assigned to each character individually. Each focuses on capturing the distinctive traits of its respective character, such as unique body shapes or specific character features.
        \item \texttt{[shared\_id]}: A common token shared across all characters. Its purpose is to capture the overall style and common characteristics of the training set, such as the color palette, textures, and outlines.
        \item \texttt{[class]}: The class descriptor, set to ``style'', guiding the learning of \texttt{[shared\_id]} by leveraging the model’s prior knowledge of artistic styles.
    \end{itemize}
    
    \vspace{6pt}\noindent2. \textit{Token Selection:} For token selection, we utilize the ranked token list provided by~\citep{2kpr2022PromptD}, which orders all tokens of length four or less by frequency of use. First, we assign the rarest token to \texttt{[shared\_id]} to ensure that the style token has minimal pre-existing associations. Then, for the assignment of tokens to individual characters we introduce two selection strategies:

    \begin{itemize}
        \item \textbf{Rarest:} The \texttt{[specific\_id]} tokens are selected by choosing the next $N$ rarest tokens from the list, where $N$ represents the number of characters in the dataset. This method aims to simplify the learning process reducing the interference with the model's prior knowledge.
        
        \item \textbf{Clustered:} In this approach, we prioritize selecting tokens that are as distinct as possible from each other within the model's learned embedding space to prevent unintended semantic overlaps that could jeopardize the distinction between characters. 
        To achieve this, we first extract the text embedding for all candidate tokens~\citep{2kpr2022PromptD} using CLIP's text encoder~\citep{radford2021LearningTVMNLS}. Then, we apply k-means clustering with cosine distance, using k-means++ initialization, 10 restarts, and a maximum of 50 iterations, partitioning the embedding space into $N$ clusters. Finally, from each cluster, we select the token closest to the centroid and assign it to the corresponding \texttt{[specific\_id]}.
    \end{itemize}
    
    \vspace{6pt}\noindent3. \textit{No class-specific regularization:}  We remove the regularization dataset intended to preserve the model’s prior knowledge, as our goal is for the model to focus exclusively on generating new characters within a specific artistic style, without concern for the potential loss of generalization across other styles.

\vspace{6pt}\noindent\textbf{Generation.} To enable the automated creation of an unlimited number of unique characters, we introduce two generation methods that eliminate the need for manually crafted prompts. Both approaches follow the same prompt structure:
\begin{center}
    ``\texttt{[token $|$ embedding] [shared\_id]} style''.
\end{center}
 Here, \texttt{[shared\_id]} represents the rare token assigned during training to capture shared visual characteristics, while \texttt{[token $|$ embedding]} refers to the randomly selected element that varies depending on the employed method. We define two such methods: 

\begin{enumerate}
    \item \textit{Random Tokens:} This method generates new characters by randomly selecting unused tokens from the same rare token list employed during training. Excluding tokens already assigned to specific characters or styles, we sample from the remaining pool to introduce variation while ensuring compatibility with the model’s learned representations.
    
    \item \textit{Random Embeddings:} Instead of relying on discrete tokens, this method generates embeddings directly, bypassing the tokenization process. The embeddings are injected into the model’s text encoder to define new character identities. We explore two variants of this method: one based on univariate Gaussian sampling, and a novel multivariate approach that leverages the covariance structure of rare token embeddings to produce more diverse and coherent character designs. In both cases, the statistical parameters (mean, standard deviation, and covariance matrix) are computed from the CLIP embeddings of rare tokens.
\end{enumerate}

The Random Token method ensures variability while utilizing existing tokens from the model's vocabulary. In contrast, the Random Embedding method extends beyond the constraints of predefined tokens, enabling a theoretically unlimited variety of character identities. However, since Stable Diffusion has limited familiarity with purely random embeddings, their effectiveness may vary, potentially producing unexpected results. All three approaches are evaluated in our experimental study (Section \ref{sec-results}).

\section{Experimental setup}\label{sec-experiment}

In this section, we present the experimental framework used to evaluate the proposed method. The datasets employed for experimentation are described in Section \ref{subsec-datasets}, followed by an explanation of the evaluation metrics in \ref{subsec-metrics}. Section \ref{subsec-baselines} details the model training process and the methods used for comparison. Finally, the experimental design is outlined in Section \ref{subsec-expD}.

\subsection{Datasets}\label{subsec-datasets}

The datasets used in this study are selected based on the nature of the problem. In general, we establish three key selection criteria:
\begin{enumerate}
    \item \textit{Few-shot constraint}. Each dataset contains only 10 to 30 images to maintain the few-shot learning setting. 
    \item \textit{Challenging textual descriptiveness}. The datasets consist of characters whose styles and visual features are difficult to fully describe using text.
    \item \textit{Consistent artistic style}. All characters within a dataset share a coherent visual style. 
\end{enumerate}
 
\begin{figure*}[!ht]
    \centering
    \resizebox{\linewidth}{!}{%
    \begin{tikzpicture}[picture format/.style={inner sep=15pt,}]
        
         \node[picture format,anchor=north] (D1) {\includegraphics[width=3in]{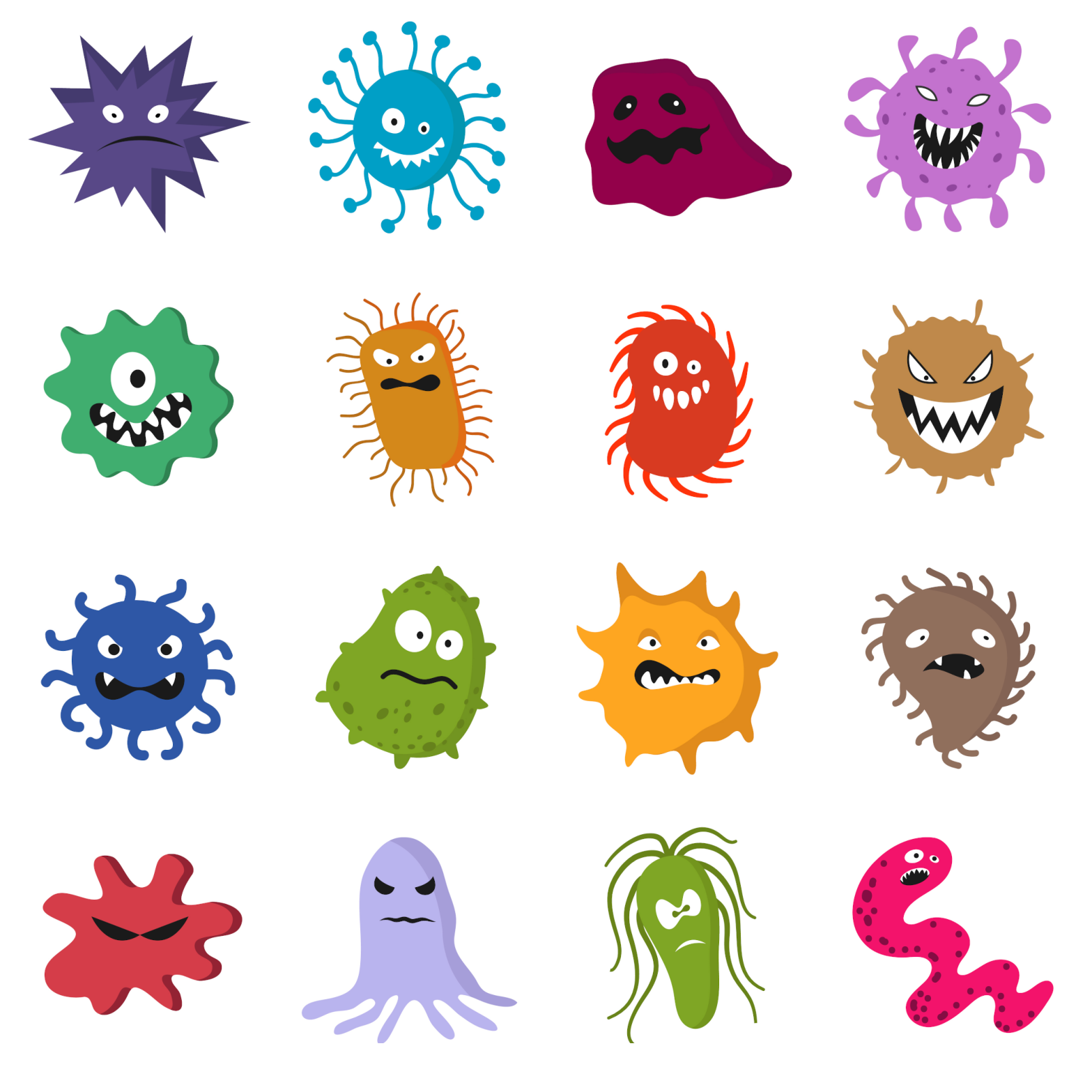}};
          \node[picture format,anchor=north west] (D2) at (D1.north east) {\includegraphics[width=3in]{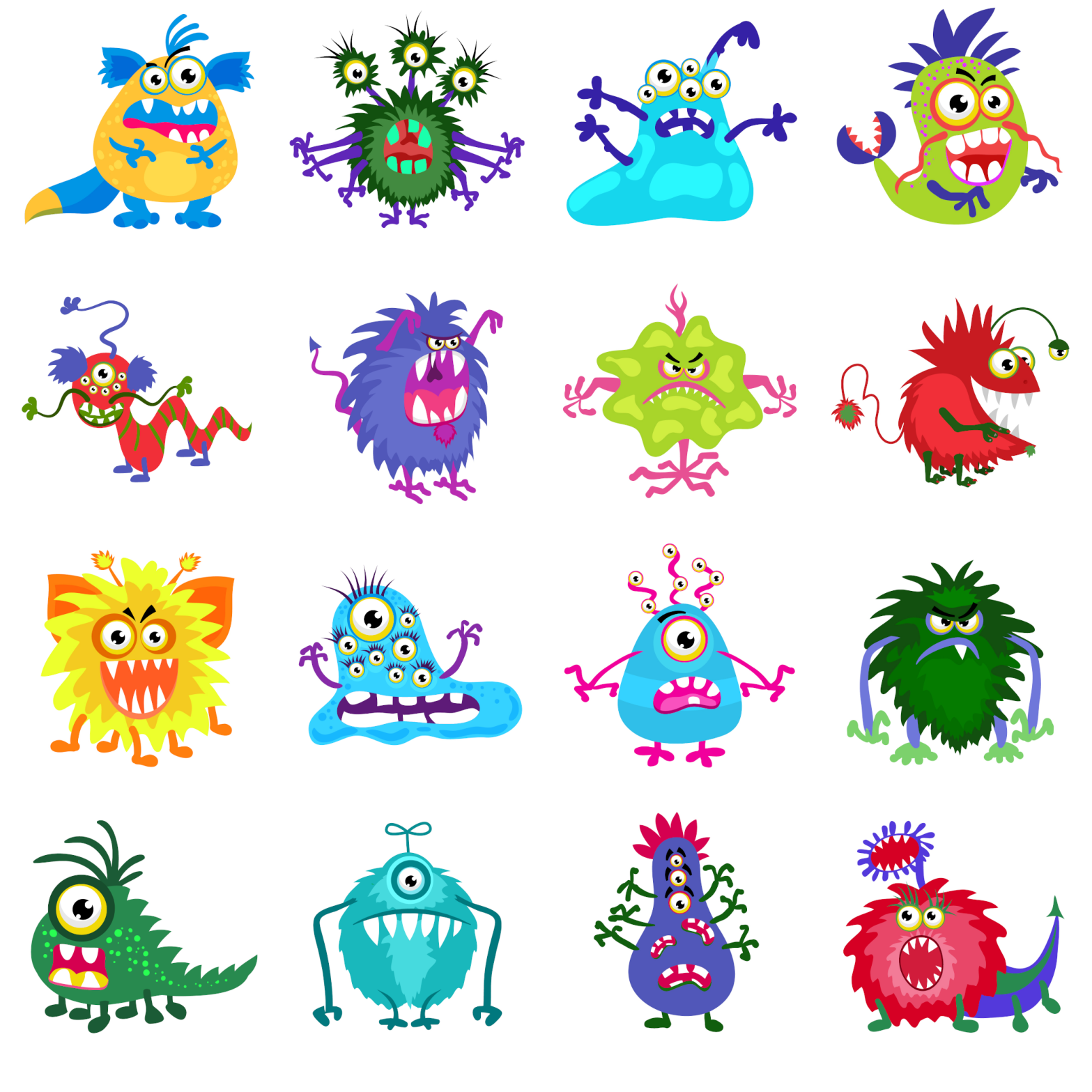}}; 
          \node[picture format,anchor=north west] (D3) at (D2.north east) {\includegraphics[width=3in]{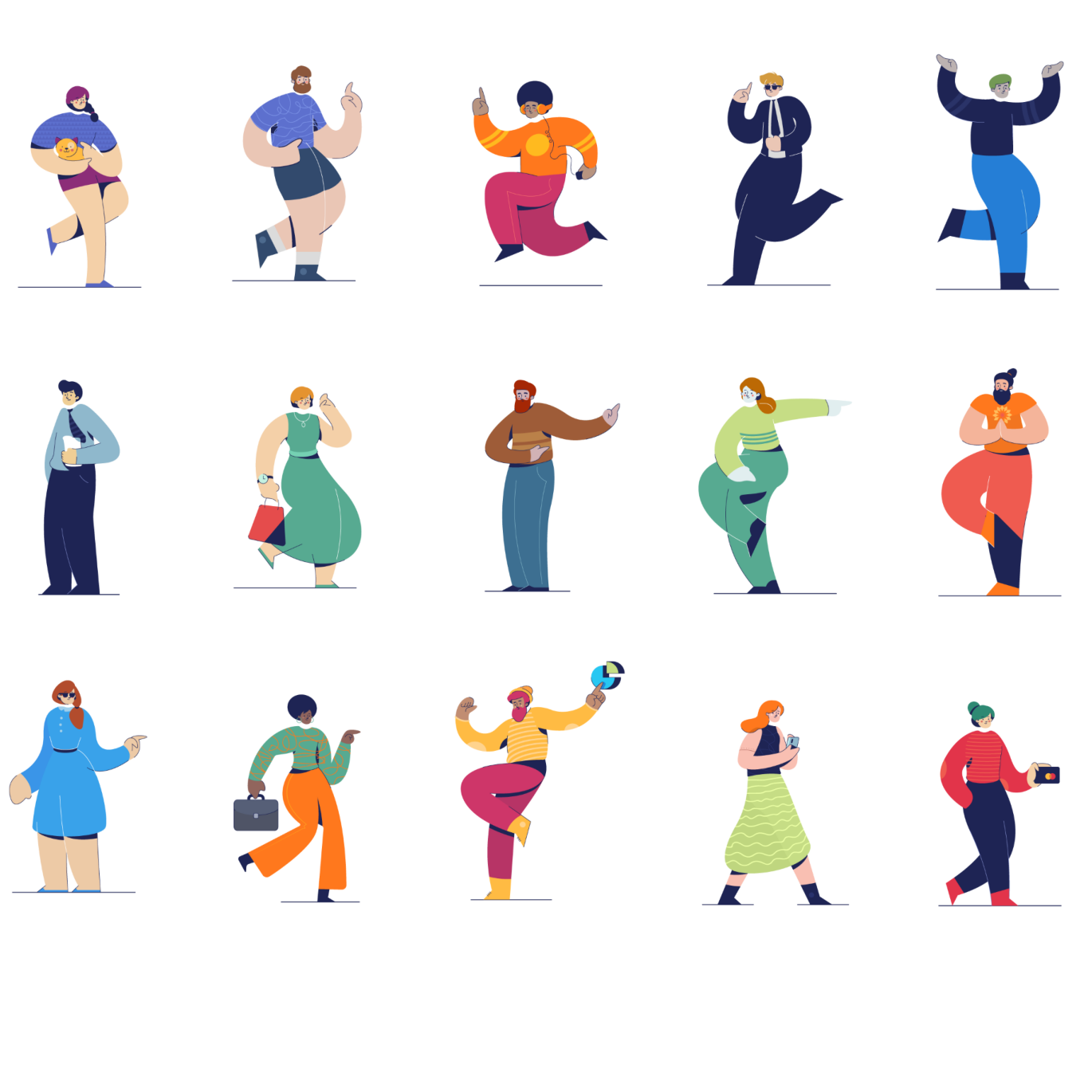}}; 
          \node[picture format,anchor=north west] (D4) at (D3.north east) {\includegraphics[width=3in]{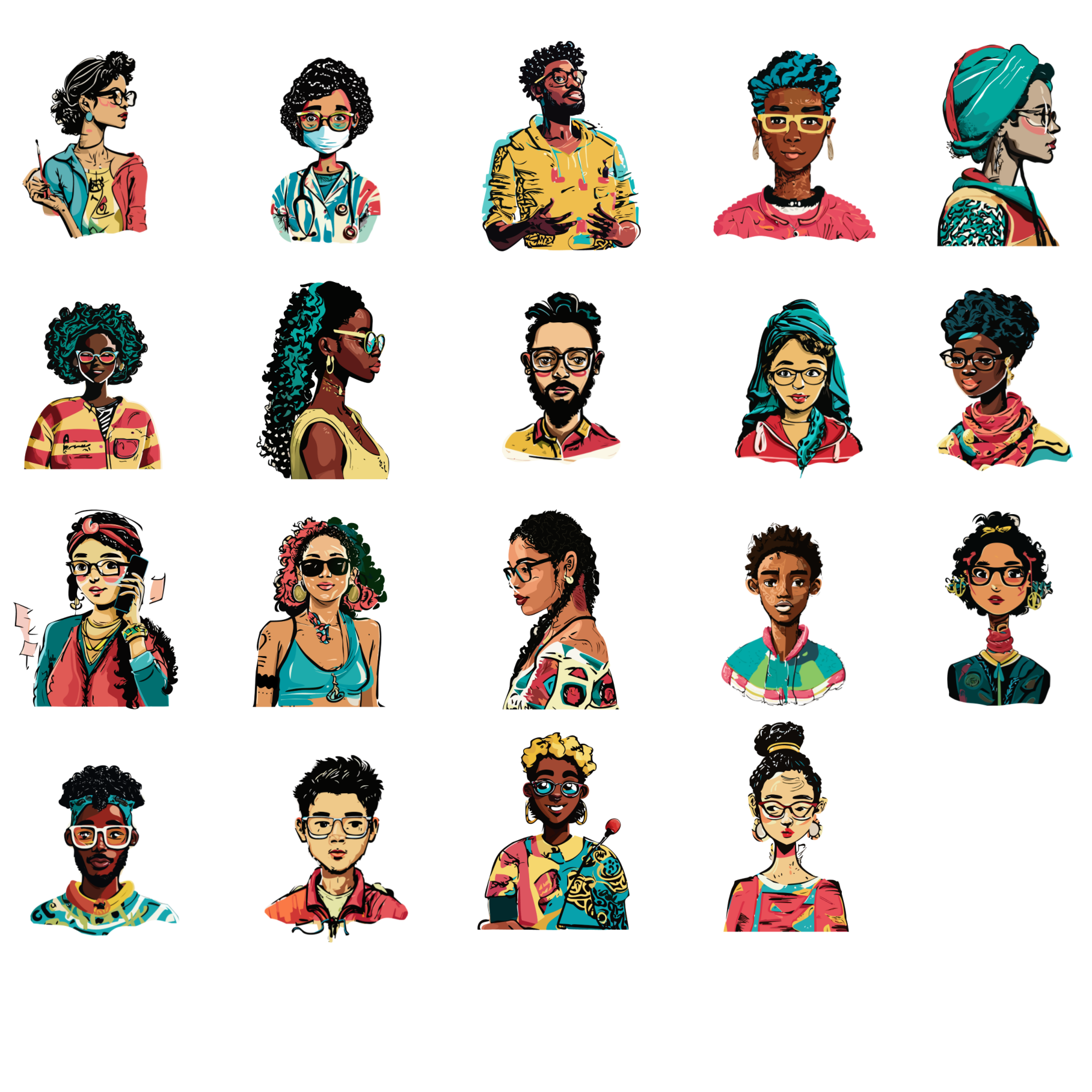}};
          \node[picture format,anchor=north west] (D5) at (D4.north east) {\includegraphics[width=4in]{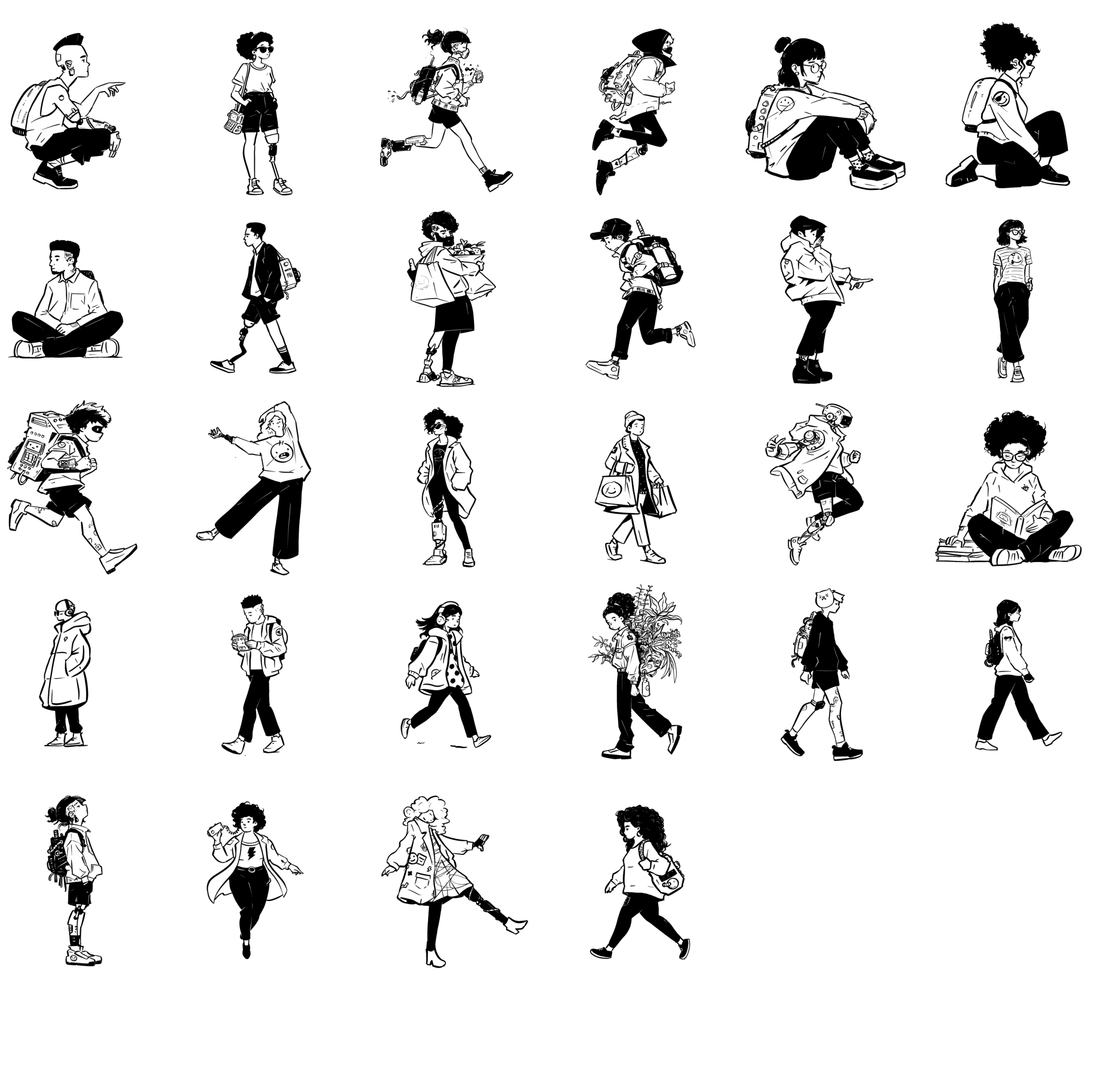}};

          \node[anchor=south] (L1) at (D1.north)  {\large (a) Virus  \citep{macrovector2023VirusD}};
          \node[anchor=south] (L2) at (D2.north)  {\large (b) Scary \citep{macrovector2023ScaryD}};
          \node[anchor=south] (L3) at (D3.north)  {\large (c) Daily \citep{getillustrations2023DailyD}};
          \node[anchor=south] (L4) at (D4.north)  {\large (d) Hipster \citep{getillustrations2023HipsterD}};
          \node[anchor=south] (L5) at (D5.north)  {\large (e) Trans \citep{stanley2023TransD}};

        \end{tikzpicture}
    }
    \caption{Character datasets used for experiments.}
    \label{fig:datasets}
\end{figure*}
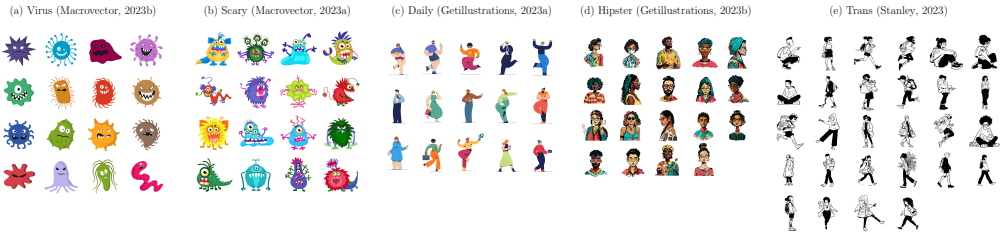

Following these criteria, we gathered five datasets (Figure \ref{fig:datasets}):

\begin{itemize}
    \item \textbf{Virus}~\citep{macrovector2023VirusD}: A collection of 16 cartoon-style virus illustrations. While the viruses exhibit some diversity, primarily through color variations, their designs remain relatively simple.
    
    \item \textbf{Scary}~\citep{macrovector2023ScaryD}: Contains 16 images of cartoon-style monsters. These characters feature complex and distinctive traits, such as multiple limbs, extra eyes, and vibrant color palettes, making them more challenging to describe accurately using text.
    
    \item \textbf{Daily}~\citep{getillustrations2023DailyD}: Includes 15 full-body cartoon-style human characters. This dataset emphasizes the overall character representation but with less focus on fine details.
    
    \item \textbf{Hipster}~\citep{getillustrations2023HipsterD}: Comprises 19 bust illustrations of cartoon-style human characters. This dataset presents a greater challenge through detailed facial features and unconventional color schemes, while still depicting a familiar subject.
    
    \item \textbf{Trans}~\citep{stanley2023TransD}: Consists of 28 full-body, line-art-style human illustrations. Unlike the other datasets, Trans features a larger sample size and greater complexity by depicting characters with prosthetics and assistive devices, as well as the absence of color. Alongside Daily and Hipster, this dataset focuses on human characters.
\end{itemize}

\begin{figure}[!ht]
    \centering
    \includegraphics[width=\linewidth]{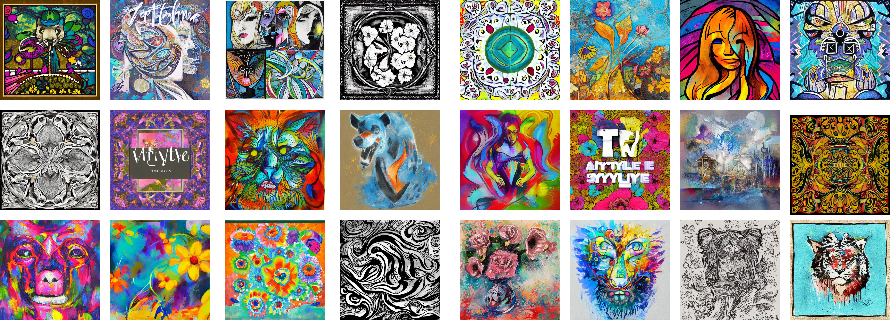}
    \caption{Example images from regularization dataset.}
    \label{fig:regdataset}
\end{figure}

In addition to the training datasets, we use a regularization dataset for training the baseline models: DreamBooth ($\text{DB}$) and DreamBooth with LoRA ($\text{DB}^{\text{L}}$) (see Section \ref{subsec-baselines}). This dataset, obtained from~\citep{nitrosocke2022RegularizationD}, consists of 1,554 images generated with Stable Diffusion v1-5.
Figure \ref{fig:regdataset} presents sample images from the regularization dataset.

\subsection{Evaluation metrics}\label{subsec-metrics}

Evaluating the generated characters presents two main challenges: no single metric effectively quantifies both style fidelity and character diversity, and quantitative metrics become less reliable at distinguishing quality when generation reaches a high standard. 

To address these issues, we assess model performance through two distinct evaluation methods. The first consists of performance metrics, which quantitatively compare generated images with the reference datasets. 
The second is human evaluation, which captures subjective aspects of visual coherence and artistic consistency that objective metrics alone may not fully reflect, particularly when models achieve similar numerical scores.

\vspace{6pt}\noindent\textbf{Quantitative metrics}. We use two quantitative metrics: Fidelity (for style preservation) and Diversity (for character variation).

\vspace{6pt}\noindent\textit{Fidelity}: To assess how closely the generated characters match the style and shared characteristics of a reference dataset, we follow the image to image CLIP similarity (CLIP-I) used in DreamBooth~\citep{ruiz2023DreamBoothFTTDMSG}, the image to image analogue of the well-known CLIPScore~\citep{hessel2021CLIPScoreREMIC}. To better capture style, we replace CLIP’s standard image encoder with a version specifically trained to distinguish visual styles~\citep{somepalli2025InvestigatingSSDM}. We refer to the resulting style-adapted CLIP-I as \textit{Fidelity}.

Given a set of $n$ generated images $V$ with visual embeddings $v_1, \ldots, v_n$ and a dataset of $m$ reference images $D$ with embeddings $d_1, \ldots, d_m$, Fidelity is computed as:

\begin{equation}
       \text{Fidelity}(V,D) = \frac{1}{nm}\sum_{i=1}^n\sum_{j=1}^m{\max(\cos(v_i,d_j), 0)}.
\end{equation}

Higher Fidelity values indicate that the generated images closely align with the visual style of the reference dataset.

\vspace{6pt}\noindent\textit{Diversity:} In addition to the Fidelity metric, we also assess the diversity of the generated characters, as greater diversity indicates a higher number of unique character designs. To this end, we adopt the metric proposed in~\citep{lai2020DiversityDHQCMTC}, originally designed for text collections but applicable to image embeddings. 

This metric operates in the same high-dimensional space as CLIP's image encoder and measures diversity by computing the standard deviation of each embedding component across all generated images. Given a set of generated images $V$, Diversity is defined as:
\begin{equation}
    \text{Diversity}(V) = \sqrt[H]{\prod_{i=1}^{H}\sigma_{i}},
\end{equation}
where $H$ is the dimensionality of the image embeddings, and $\sigma_i$ represents the standard deviation of the $i$-th component. A higher standard deviation indicates greater variation in the generated characters, reflecting a broader range of unique designs.

\vspace{6pt}\noindent\textbf{Human evaluation}.
To compare the studied methods beyond what quantitative metrics can capture, we follow best practices for evaluating generative models~\citep{otani2023VerifiableRHETG} and conduct a human evaluation using two complementary methods: pairwise comparisons and absolute ratings.

\vspace{6pt}\noindent\textit{Comparison-based evaluation}: In this approach, participants are shown pairs of images generated by different methods and asked to select which one better fits the reference style, or indicate a tie. We use the Bradley-Terry (BT) model~\citep{bradley1952RankAIBDMPCa} to aggregate comparisons into a global ranking. 

\vspace{6pt}\noindent\textit{Rating-based evaluation}: In this approach, participants rate single images on a fixed scale based on how well each image fits the reference set. Scores are averaged across evaluators to obtain a global score per method. This approach enables more detailed feedback and is well-suited for expert participants familiar with visual design and style evaluation~\citep{ioannou2024EvaluationNSTR}.

This evaluation strategy is motivated by the different expertise levels of the participant groups. Pairwise comparisons are assigned to general participants, as this method is known to improve reliability when used by non-experts~\citep{hoeijmakers2024HowSCIQABSRPCILS}, and is widely adopted in adjacent fields such as large language model evaluation in Chatbot Arena LLM~\citep{chiang2024ChatbotAOPELHP}. In contrast, absolute ratings are reserved for professional artists, who are better equipped to make calibrated judgments about stylistic fidelity~\citep{otani2023VerifiableRHETG}.

\subsection{Comparison methods and training setup}\label{subsec-baselines}

Considering that our training method builds upon DreamBooth ($\text{DB}$), we use it as the primary baseline. We further include Textual Inversion ($\text{TI}$), a widely adopted personalization approach, for comparison. Finally, since we propose the integration of LoRA with DreamBooth to mitigate overfitting, we also consider the basic application of LoRA with DreamBooth as a secondary baseline ($\text{DB}^{\text{L}}$). 

$\text{TI}$ is trained following the procedure described in the original Textual Inversion paper~\citep{gal2022ImageWOWPTGUTI}, whereas $\text{DB}$ and $\text{DB}^{\text{L}}$ adhere to the process outlined in~\citep{ruiz2023DreamBoothFTTDMSG} and make use of the same regularization dataset (see Section \ref{subsec-datasets}). To maintain consistency, a randomly selected subset of regularization images is used in each epoch, ensuring that the number of selected images matches the size of the training dataset.  

Beyond these baselines, we evaluate two versions of our proposed approach, which follow the same training procedure but differ in their token selection strategy. Specifically, we refer to the version that employs rarest selection as $\text{DB}_{\text{MT-R}}^{\text{L}}$ and the one that uses clustered selection as $\text{DB}_{\text{MT-C}}^{\text{L}}$.  

To ensure a robust evaluation, we train five independent instances of each training method on each dataset, resulting in a total of 125 models ($5 \ \text{training methods} \times 5 \ \text{copies} \times 5 \ \text{datasets}$). Using these models, we generate images with three different generation methods (random token, univariate random embedding and multivariate random embedding), producing 400 images per model for a total of 150,000 images ($125 \ \text{models} \times 3 \ \text{generation methods} \times 400 \ \text{images}$). 

To further analyze our design choices, we also conduct an ablation study (Appendix~\ref{app1}) under the same training setup. This study isolates the effect of the class-specific regularization set and of multi-token training in $\text{DB}^{\text{L}}_{\text{MT-C}}$, providing insight into how each factor impacts validity, style fidelity, and diversity.

All methods are implemented in PyTorch 2.2.0 and trained with identical hyperparameters: Stable Diffusion v1-5 as the base model, 20,000 training steps, an image resolution of $512 \times 512$ pixels, a batch size of 4, a learning rate of $10^{-4}$, a cosine noise scheduler, a caption dropout rate of 0.25, and a noise offset of 0.1. Training is conducted on a machine equipped with an NVIDIA 3080 Ti, an Intel Core i5-11400 processor, and 32 GB of RAM.

\subsection{Evaluation procedure}\label{subsec-expD}

To evaluate the results of the different methods, we adopt three complementary evaluation approaches.

\subsubsection{Approach 1. Validity of generated images}\label{subsubsec-filter} Before applying the quantitative metrics described in Section~\ref{subsec-metrics}, we first assess the validity of the generated images. This step serves to evaluate the overall quality of the generations by measuring the frequency with which each method produces outputs that are invalid, defective, or inconsistent with the intended style or structure. We define four primary types of invalid outputs:

\begin{enumerate}
    \item \textit{Training copies}: Some images are exact or near copies of the original training images. To detect these, we compute the CLIP-I similarity between each generated image and the closest reference image. Images exceeding a defined similarity threshold are invalidated. This threshold was selected individually for each dataset through visual inspection to reliably identify near-identical cases.
    \item \textit{Defective images}: Some images fail to align with the intended style or subject, either deviating visually from the dataset’s artistic characteristics or generating content that does not match the expected category. To detect these, we compute the Fidelity between each generated image and all images in the dataset. Images whose Fidelity falls below a dataset-specific threshold are marked as defective. This threshold was also determined individually for each dataset by visual inspection to capture incoherent generations.
    \item \textit{Multiple subjects}: Some images contain multiple subjects when only one is expected (e.g., two characters instead of one). We detect these cases using the segment anything model (SAM)~\citep{kirillov2023SegmentA}, followed with a manual filtering process to correct possible errors. A human evaluator reviews flagged images to reduce false positives.
    \item \textit{Duplicate outputs}: Certain models produce identical images across multiple generations, suggesting overfitting or mode collapse. We detect duplicates using the Structural Similarity Index (SSIM)~\citep{wang2004ImageQAEVSS}, retaining only one valid instance while marking the rest as invalid.
\end{enumerate}

The filtering process is applied sequentially in the order above. If an image is invalidated by one criterion, it is excluded from further evaluation. 

\subsubsection{Approach 2. Quantitative performance comparison}\label{subsubsec-objEval}

The goal of this study is to quantitatively compare the performance of the studied methods in terms of stylistic fidelity and character diversity. We apply the metrics introduced in Section~\ref{subsec-metrics}--Fidelity and Diversity--to the valid images obtained after Approach 1. Results are computed per dataset, generation method, and training instance, and are reported as the mean and standard deviation across the five independently trained models for each method. These results also guide the selection of methods for the human evaluation described in Approach 3.

\subsubsection{Approach 3. Human Evaluation}\label{subsubsec-humanEval}

To further evaluate the established methods, we conduct a human evaluation with two participant groups: general participants (135 university students, aged 18–23) and professional artists (6 participants). 
Throughout the evaluation, four reference images from the dataset are displayed as a consistent visual guide. Each participant evaluates all datasets and methods. For each dataset, participants are shown a random selection of generated images. Additionally, to serve as a control condition, some images are drawn from the dataset itself and treated as if they were generated by an additional method (always distinct from the four reference images). 

General participants completed six pairwise comparisons per dataset, resulting in up to 30 comparisons per participant. Due to partial completion by some participants, the total number of pairwise comparisons collected was 3,920. Professional artists were shown 10 individual images per dataset, resulting in 300 ratings.

As mentioned in Section \ref{subsec-metrics}, the evaluation procedure differs between groups to account for their level of expertise, as described below.

    \vspace{6pt}\noindent\textbf{General participants} perform a comparative evaluation, where they are presented with two images generated by different methods and asked:
    \begin{quote}
        ``Which of the two characters fits better as a new member of the family (in terms of the style and characteristics)?''
        
        Response options: ``A is better", ``B is better'', ``Both fit equally'', and ``Neither fits''.
    \end{quote}

    \vspace{6pt}\noindent\textbf{Professional artists} perform a rating-based evaluation, where they are presented with a single image generated by one of the methods and rate how well it integrates with the reference characters. This is the given instruction:
    \begin{quote}
        ``Rate how well the following image integrates as part of the presented family.''

        Scale: 1 (does not fit at all) -- 5 (could pass as another family character).
    \end{quote}

\section{Results}\label{sec-results}

In this section, we report the outcomes of the three studies described in Section \ref{subsec-expD}. We begin by presenting the validity statistics of the generated images across methods and datasets (Section \ref{subsec-filter}), followed by quantitative results based on the Fidelity and Diversity metrics (Section \ref{subsec-comparison}). Finally, we report the results of the human evaluation (Section \ref{subsec-humanEval}). 

\subsection{Approach 1.Validity of generated images}\label{subsec-filter}

\begin{figure*}[htp]
    \centering
    \resizebox{\textwidth}{!}{%
        \includegraphics[]{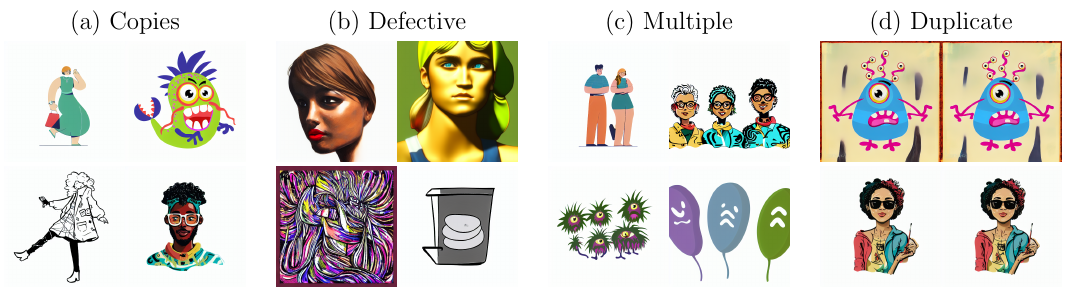}
    }
    \caption{Examples of generated images that fail to meet the validity criteria, classified by the corresponding reasons for their rejection.}
    \label{fig:invExamples}
\end{figure*}

Table~\ref{tab:exp1InvVal} provides a breakdown of percentage of invalid images for each dataset and training method combination, categorized by specific reason for invalidity. The Total column summarizes the total percentage of invalid images for each combination. Figure~\ref{fig:invExamples} shows representative examples of invalid outputs for each category.

\begin{table}[ht!]
\centering
\caption{Summary of invalid images for each dataset and training method combination, categorized by the specific invalidity reason. Cases where invalid images exceed 5\% of the total generated images are highlighted in \textbf{bold}.}
\resizebox{0.8\linewidth}{!}{
\def\arraystretch{1.2}
\begin{tabular}{llrrrrr}
\toprule
Dataset & Training & Copies & Defective & Multiple & Duplicate & Total \\
\midrule
\multirow[t]{5}{*}{Virus} & $TI$ & $0.00\%$ & $\pmb{53.03\%}$ & $3.42\%$ & $0.00\%$ & $\pmb{56.45\%}$ \\
  & $DB$ & $\pmb{88.97\%}$ & $2.87\%$ & $0.00\%$ & $\pmb{7.78\%}$ & $\pmb{99.62\%}$ \\
  & $DB^{{L}}$ & $0.00\%$ & $\pmb{6.27\%}$ & $2.43\%$ & $0.00\%$ & $\pmb{8.70\%}$ \\
  & $DB_{{MT-R}}^{{L}}$ & $0.02\%$ & $0.17\%$ & $0.00\%$ & $0.00\%$ & $0.18\%$ \\
  & $DB_{{MT-C}}^{{L}}$ & $0.93\%$ & $1.03\%$ & $0.00\%$ & $0.10\%$ & $2.07\%$ \\
\midrule
\multirow[t]{5}{*}{Scary} & $TI$ & $0.00\%$ & $\pmb{61.22\%}$ & $1.28\%$ & $0.00\%$ & $\pmb{62.50\%}$ \\
  & $DB$ & $\pmb{92.62\%}$ & $\pmb{7.22\%}$ & $0.00\%$ & $0.05\%$ & $\pmb{99.88\%}$ \\
  & $DB^{{L}}$ & $0.00\%$ & $0.85\%$ & $1.00\%$ & $0.00\%$ & $1.85\%$ \\
  & $DB_{{MT-R}}^{{L}}$ & $0.03\%$ & $3.28\%$ & $0.00\%$ & $0.00\%$ & $3.32\%$ \\
  & $DB_{{MT-C}}^{{L}}$ & $0.05\%$ & $0.00\%$ & $0.03\%$ & $0.00\%$ & $0.08\%$ \\
\midrule
\multirow[t]{5}{*}{Daily} & $TI$ & $0.00\%$ & $\pmb{83.82\%}$ & $\pmb{6.20\%}$ & $0.00\%$ & $\pmb{90.02\%}$ \\
  & $DB$ & $\pmb{100.00\%}$ & $0.00\%$ & $0.00\%$ & $0.00\%$ & $\pmb{100.00\%}$ \\
  & $DB^{{L}}$ & $0.00\%$ & $0.55\%$ & $1.23\%$ & $0.00\%$ & $1.78\%$ \\
  & $DB_{{MT-R}}^{{L}}$ & $0.02\%$ & $0.00\%$ & $3.53\%$ & $0.00\%$ & $3.55\%$ \\
  & $DB_{{MT-C}}^{{L}}$ & $0.02\%$ & $0.00\%$ & $2.23\%$ & $0.00\%$ & $2.25\%$ \\
\midrule
\multirow[t]{5}{*}{Hipster} & $TI$ & $0.00\%$ & $\pmb{86.37\%}$ & $3.30\%$ & $0.00\%$ & $\pmb{89.67\%}$ \\
  & $DB$ & $\pmb{99.60\%}$ & $0.07\%$ & $0.00\%$ & $0.20\%$ & $\pmb{99.87\%}$ \\
  & $DB^{{L}}$ & $0.07\%$ & $\pmb{12.47\%}$ & $\pmb{49.52\%}$ & $0.00\%$ & $\pmb{62.05\%}$ \\
  & $DB_{{MT-R}}^{{L}}$ & $0.35\%$ & $2.22\%$ & $1.72\%$ & $0.00\%$ & $4.28\%$ \\
  & $DB_{{MT-C}}^{{L}}$ & $0.15\%$ & $1.27\%$ & $2.08\%$ & $0.00\%$ & $3.50\%$ \\
\midrule
\multirow[t]{5}{*}{Trans} & $TI$ & $0.00\%$ & $\pmb{76.73\%}$ & $3.02\%$ & $0.00\%$ & $\pmb{79.75\%}$ \\
  & $DB$ & $\pmb{87.00\%}$ & $\pmb{12.82\%}$ & $0.02\%$ & $0.03\%$ & $\pmb{99.87\%}$ \\
  & $DB^{{L}}$ & $0.12\%$ & $\pmb{5.15\%}$ & $1.57\%$ & $0.00\%$ & $\pmb{6.83\%}$ \\
  & $DB_{{MT-R}}^{{L}}$ & $0.43\%$ & $0.30\%$ & $0.92\%$ & $0.00\%$ & $1.65\%$ \\
  & $DB_{{MT-C}}^{{L}}$ & $0.10\%$ & $0.13\%$ & $0.90\%$ & $0.00\%$ & $1.13\%$ \\
\bottomrule
\end{tabular}
}
\label{tab:exp1InvVal}
\end{table}

Across all datasets, $\text{DB}$ produces the highest number of invalid images, primarily classified under the \textit{Training copies} category. $\text{TI}$ also yields a large proportion of invalid samples, most frequently in the \textit{Defective} category, making it the second weakest method in terms of validity. 

The $\text{DB}^{\text{L}}$ method results in fewer invalid images overall, though defective images are more frequent, particularly in the Virus, Hipster, and Trans datasets. In the Hipster dataset, the \textit{Multiple subjects} category account for a substantial portion of invalid images generated by $\text{DB}^{\text{L}}$.

The proposed methods, $\text{DB}_{\text{MT-R}}^{\text{L}}$ and $\text{DB}_{\text{MT-C}}^{\text{L}}$, generate the lowest number of invalid images across all datasets. Lastly, the issue of \textit{duplicate outputs} is rare overall, with only one notable occurrence in the Virus dataset under the baseline $\text{DB}$ method.

\subsection{Approach 2. Quantitative performance comparison}\label{subsec-comparison}

\begin{figure*}[htp]
    \centering
    \resizebox{\textwidth}{!}{
        \includegraphics[]{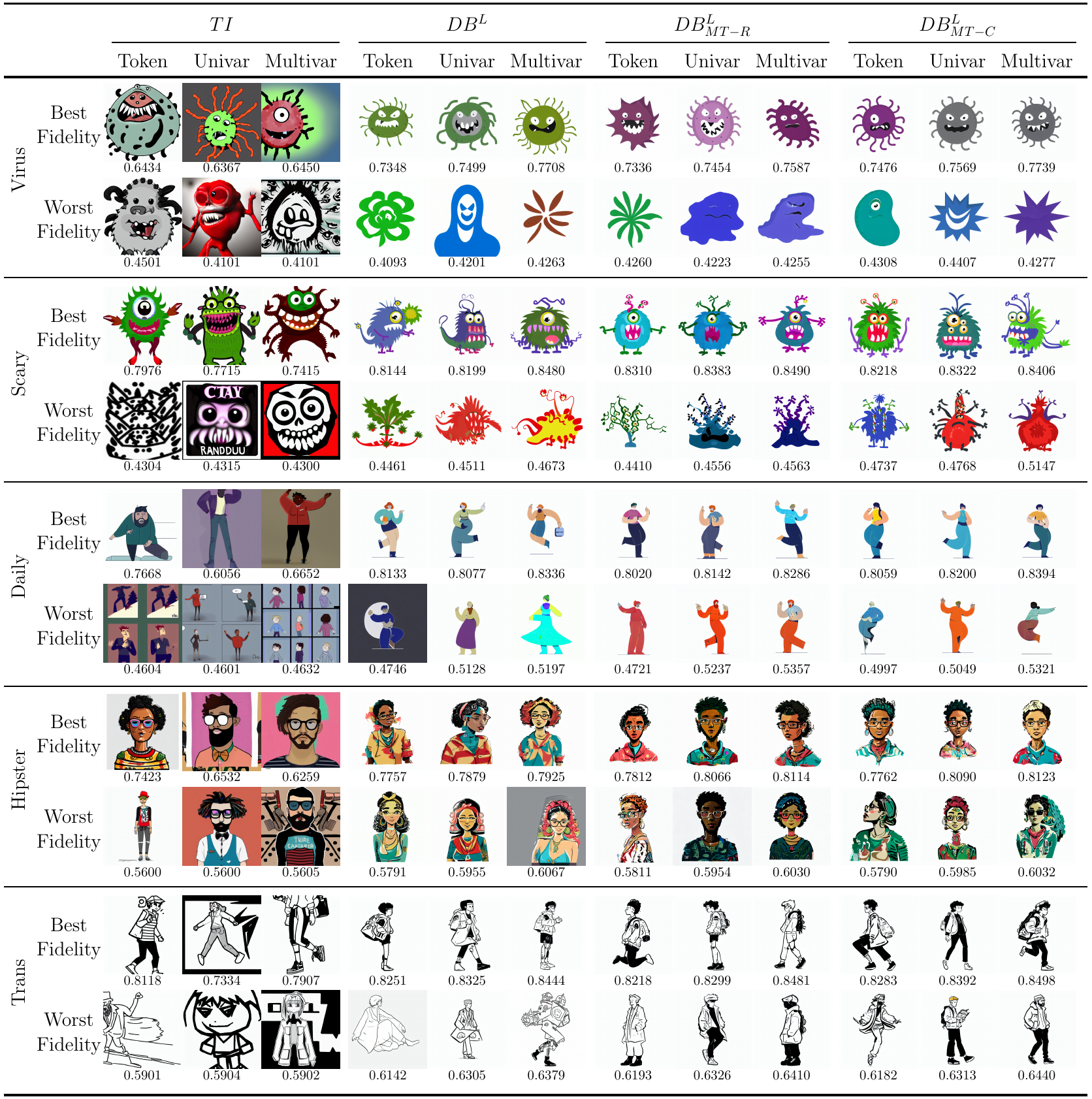}
    }
    \caption{Examples of generated images with the highest and lowest Fidelity scores for each dataset, model, and generation method. Fidelity scores are shown below each image.}
    \label{fig:Examples}
\end{figure*}

Table \ref{tab:exp1Res} reports the average Fidelity and Diversity scores, along with standard deviations. Metrics are computed using all valid images obtained after filtering in Approach 1. Combinations where more than 95\% of the generated images are invalid are excluded from the comparison, as such a high invalidity rate indicates poor model performance, making these cases unsuitable for meaningful comparison. These exclusions are marked with a ``-'' in the table. Additionally, Figure~\ref{fig:Examples} presents examples of generated images with the highest and lowest Fidelity scores, illustrating qualitative differences corresponding to these quantitative results. All generated images are included in the supplementary materials, along with the Fidelity and Diversity values reported for each dataset, and individual Fidelity scores for each generated image.

\begin{sidewaystable}
\centering
\caption{Fidelity and Diversity metric results for valid images generated using each combination of training and generation methods across the five presented datasets. The invalid image count is included to provide context on the generation capabilities of each combination. Each row represents a specific training and generation method pairing. The best values for each dataset and metric are highlighted in \textbf{bold}.}
\adjustbox{width=\textwidth, center}{
\begin{tabular}{llrrrrrrrrrrrrrrr}
\toprule
 &  & \multicolumn{3}{l}{Virus} & \multicolumn{3}{l}{Scary} & \multicolumn{3}{l}{Daily} & \multicolumn{3}{l}{Hipster} & \multicolumn{3}{l}{Trans} \\
 \cmidrule(r){3-5} \cmidrule(r){6-8} \cmidrule(r){9-11} \cmidrule(r){12-14} \cmidrule(r){15-17}
 Training & Generation  & Invalid ↓ & Fidelity ↑ & Diversity ↑ & Invalid ↓ & Fidelity ↑ & Diversity ↑ & Invalid ↓ & Fidelity ↑ & Diversity ↑ & Invalid ↓ & Fidelity ↑ & Diversity ↑ & Invalid ↓ & Fidelity ↑ & Diversity ↑ \\
\midrule
\multirow[t]{3}{*}{$TI$} & Token & $ 55.60\%$ & $ .4846 {\scriptscriptstyle \pm .0193}$ & $\pmb{.3524 {\scriptscriptstyle \pm .0141}}$ & $ 45.70\%$ & $ .5903 {\scriptscriptstyle \pm .0942}$ & $\pmb{.3462 {\scriptscriptstyle \pm .0550}}$ & $ 81.85\%$ & $ .5245 {\scriptscriptstyle \pm .0438}$ & $\pmb{.3585 {\scriptscriptstyle \pm .0113}}$ & $ 74.25\%$ & $ .6186 {\scriptscriptstyle \pm .0119}$ & $\pmb{.3437 {\scriptscriptstyle \pm .0140}}$ & $ 63.40\%$ & $ .6549 {\scriptscriptstyle \pm .0202}$ & $ .3097 {\scriptscriptstyle \pm .0872}$ \\
 & Univar & $ 59.40\%$ & $ .3871 {\scriptscriptstyle \pm .2168}$ & $ .2772 {\scriptscriptstyle \pm .1558}$ & $ 70.05\%$ & $ .5005 {\scriptscriptstyle \pm .0406}$ & $ .3312 {\scriptscriptstyle \pm .0493}$ & $ 95.15\%$ & - & - & $ 96.60\%$ & - & - & $ 92.05\%$ & $ .3806 {\scriptscriptstyle \pm .3474}$ & $ .1878 {\scriptscriptstyle \pm .1716}$ \\
 & Multivar & $ 54.35\%$ & $ .4726 {\scriptscriptstyle \pm .0395}$ & $ .2752 {\scriptscriptstyle \pm .1545}$ & $ 71.75\%$ & $ .4915 {\scriptscriptstyle \pm .0497}$ & $ .2660 {\scriptscriptstyle \pm .1537}$ & $ 93.05\%$ & $ .3963 {\scriptscriptstyle \pm .2228}$ & $ .2059 {\scriptscriptstyle \pm .1881}$ & $ 98.15\%$ & - & - & $ 83.80\%$ & $ .3831 {\scriptscriptstyle \pm .3498}$ & $ .1955 {\scriptscriptstyle \pm .1787}$ \\
\midrule
\multirow[t]{3}{*}{$DB$} & Token & $ 99.60\%$ & - & - & $ 99.95\%$ & - & - & $ 100.00\%$ & - & - & $ 99.60\%$ & - & - & $ 99.85\%$ & - & - \\
 & Univar & $ 99.75\%$ & - & - & $ 99.85\%$ & - & - & $ 100.00\%$ & - & - & $ 100.00\%$ & - & - & $ 99.95\%$ & - & - \\
 & Multivar & $ 99.50\%$ & - & - & $ 99.85\%$ & - & - & $ 100.00\%$ & - & - & $ 100.00\%$ & - & - & $ 99.80\%$ & - & - \\
\midrule
\multirow[t]{3}{*}{$DB^{{L}}$} & Token & $ 13.10\%$ & $ .6184 {\scriptscriptstyle \pm .0105}$ & $ .3480 {\scriptscriptstyle \pm .0075}$ & $ 3.00\%$ & $ .7241 {\scriptscriptstyle \pm .0066}$ & $ .2945 {\scriptscriptstyle \pm .0061}$ & $ 2.75\%$ & $ .7128 {\scriptscriptstyle \pm .0035}$ & $ .2741 {\scriptscriptstyle \pm .0041}$ & $ 55.75\%$ & $ .7038 {\scriptscriptstyle \pm .0061}$ & $ .3077 {\scriptscriptstyle \pm .0022}$ & $ 10.25\%$ & $ .7372 {\scriptscriptstyle \pm .0017}$ & $\pmb{.3166 {\scriptscriptstyle \pm .0061}}$ \\
 & Univar & $ 7.30\%$ & $ .6001 {\scriptscriptstyle \pm .0264}$ & $ .3422 {\scriptscriptstyle \pm .0122}$ & $ 0.80\%$ & $ .6894 {\scriptscriptstyle \pm .0195}$ & $ .2882 {\scriptscriptstyle \pm .0072}$ & $ 1.35\%$ & $ .7019 {\scriptscriptstyle \pm .0020}$ & $ .2736 {\scriptscriptstyle \pm .0068}$ & $ 64.50\%$ & $ .6969 {\scriptscriptstyle \pm .0088}$ & $ .2993 {\scriptscriptstyle \pm .0076}$ & $ 4.45\%$ & $ .7349 {\scriptscriptstyle \pm .0050}$ & $ .3056 {\scriptscriptstyle \pm .0016}$ \\
 & Multivar & $ 5.70\%$ & $ .6144 {\scriptscriptstyle \pm .0146}$ & $ .3437 {\scriptscriptstyle \pm .0095}$ & $ 1.75\%$ & $ .7097 {\scriptscriptstyle \pm .0129}$ & $ .2896 {\scriptscriptstyle \pm .0050}$ & $ 1.25\%$ & $ .6996 {\scriptscriptstyle \pm .0044}$ & $ .2730 {\scriptscriptstyle \pm .0078}$ & $ 60.10\%$ & $ .7035 {\scriptscriptstyle \pm .0027}$ & $ .2953 {\scriptscriptstyle \pm .0013}$ & $ 5.80\%$ & $ .7448 {\scriptscriptstyle \pm .0020}$ & $ .3041 {\scriptscriptstyle \pm .0016}$ \\
\midrule
\multirow[t]{3}{*}{$DB_{{MT-R}}^{{L}}$} & Token & $\pmb{0.10\%}$ & $ .6573 {\scriptscriptstyle \pm .0018}$ & $ .2737 {\scriptscriptstyle \pm .0009}$ & $ 0.15\%$ & $ .7503 {\scriptscriptstyle \pm .0025}$ & $ .2574 {\scriptscriptstyle \pm .0012}$ & $ 8.65\%$ & $ .6950 {\scriptscriptstyle \pm .0032}$ & $ .2587 {\scriptscriptstyle \pm .0010}$ & $ 5.90\%$ & $ .6947 {\scriptscriptstyle \pm .0044}$ & $ .2804 {\scriptscriptstyle \pm .0021}$ & $ 4.15\%$ & $ .7511 {\scriptscriptstyle \pm .0033}$ & $ .3131 {\scriptscriptstyle \pm .0029}$ \\
 & Univar & $\pmb{0.10\%}$ & $ .6073 {\scriptscriptstyle \pm .0287}$ & $ .2745 {\scriptscriptstyle \pm .0037}$ & $ 1.55\%$ & $ .6920 {\scriptscriptstyle \pm .0364}$ & $ .2624 {\scriptscriptstyle \pm .0089}$ & $ 0.90\%$ & $ .7085 {\scriptscriptstyle \pm .0121}$ & $ .2414 {\scriptscriptstyle \pm .0067}$ & $ 3.70\%$ & $ .6896 {\scriptscriptstyle \pm .0104}$ & $ .2412 {\scriptscriptstyle \pm .0169}$ & $\pmb{0.15\%}$ & $\pmb{.7645 {\scriptscriptstyle \pm .0047}}$ & $ .2606 {\scriptscriptstyle \pm .0237}$ \\
 & Multivar & $ 0.35\%$ & $ .5956 {\scriptscriptstyle \pm .0402}$ & $ .2600 {\scriptscriptstyle \pm .0021}$ & $ 8.25\%$ & $ .6508 {\scriptscriptstyle \pm .0573}$ & $ .2680 {\scriptscriptstyle \pm .0116}$ & $ 1.10\%$ & $ .6869 {\scriptscriptstyle \pm .0039}$ & $ .2413 {\scriptscriptstyle \pm .0082}$ & $ 3.25\%$ & $ .6935 {\scriptscriptstyle \pm .0057}$ & $ .2429 {\scriptscriptstyle \pm .0188}$ & $ 0.65\%$ & $ .7586 {\scriptscriptstyle \pm .0042}$ & $ .2590 {\scriptscriptstyle \pm .0257}$ \\
\midrule
\multirow[t]{3}{*}{$DB_{{MT-C}}^{{L}}$} & Token & $ 1.60\%$ & $ .6721 {\scriptscriptstyle \pm .0025}$ & $ .2846 {\scriptscriptstyle \pm .0013}$ & $ 0.20\%$ & $\pmb{.7617 {\scriptscriptstyle \pm .0014}}$ & $ .2473 {\scriptscriptstyle \pm .0010}$ & $ 1.35\%$ & $ .7149 {\scriptscriptstyle \pm .0025}$ & $ .2583 {\scriptscriptstyle \pm .0018}$ & $ 8.75\%$ & $ .7032 {\scriptscriptstyle \pm .0057}$ & $ .2870 {\scriptscriptstyle \pm .0015}$ & $ 2.55\%$ & $ .7541 {\scriptscriptstyle \pm .0056}$ & $ .3144 {\scriptscriptstyle \pm .0015}$ \\
 & Univar & $ 1.65\%$ & $\pmb{.6859 {\scriptscriptstyle \pm .0028}}$ & $ .2274 {\scriptscriptstyle \pm .0302}$ & $\pmb{0.00\%}$ & $ .7600 {\scriptscriptstyle \pm .0020}$ & $ .2214 {\scriptscriptstyle \pm .0146}$ & $ 5.35\%$ & $\pmb{.7331 {\scriptscriptstyle \pm .0031}}$ & $ .2528 {\scriptscriptstyle \pm .0030}$ & $ 1.00\%$ & $ .6944 {\scriptscriptstyle \pm .0102}$ & $ .2507 {\scriptscriptstyle \pm .0150}$ & $ 0.40\%$ & $ .7464 {\scriptscriptstyle \pm .0055}$ & $ .2945 {\scriptscriptstyle \pm .0077}$ \\
 & Multivar & $ 2.95\%$ & $ .6201 {\scriptscriptstyle \pm .0316}$ & $ .2737 {\scriptscriptstyle \pm .0065}$ & $ 0.05\%$ & $ .7596 {\scriptscriptstyle \pm .0057}$ & $ .2276 {\scriptscriptstyle \pm .0122}$ & $\pmb{0.05\%}$ & $ .7271 {\scriptscriptstyle \pm .0047}$ & $ .2419 {\scriptscriptstyle \pm .0096}$ & $\pmb{0.75\%}$ & $\pmb{.7041 {\scriptscriptstyle \pm .0030}}$ & $ .2662 {\scriptscriptstyle \pm .0115}$ & $ 0.45\%$ & $\pmb{.7645 {\scriptscriptstyle \pm .0061}}$ & $ .2860 {\scriptscriptstyle \pm .0141}$ \\
\bottomrule
\end{tabular}
}
\label{tab:exp1Res}
\end{sidewaystable}

The first thing that stands out looking at Table \ref{tab:exp1Res} is that $\text{DB}$ baseline method produced insufficient valid outputs to compute meaningful Fidelity and Diversity metrics and is therefore excluded from further analysis. Turning to $\text{TI}$, this method is able to generate a larger number of valid samples than $\text{DB}$, but still produces a substantial proportion of invalid images, in most cases exceeding half of the generated images. Among the valid generations, $\text{TI}$ consistently exhibits lower Fidelity compared to the other methods, indicating weaker alignment with the training data. It also shows an unstable behavior in terms of Diversity: while high values are observed under the Token generation strategy, the Univar and Multivar settings yield much lower scores. 

Observing Table~\ref{tab:exp1Res}, the $\text{DB}^{\text{L}}$ method achieves moderate Fidelity values across datasets, with the highest performance observed using the Multivar generation strategy, notably on the Trans dataset ($0.7448$). Diversity values for $\text{DB}^{\text{L}}$ are the highest among the $\text{DB}$-based methods, particularly with the Token generation approach, which provides strong results across all datasets.

The proposed multi-token methods ($\text{DB}_{\text{MT-R}}^{\text{L}}$ and $\text{DB}_{\text{MT-C}}^{\text{L}}$) yield higher Fidelity scores overall. Specifically, $\text{DB}_{\text{MT-C}}^{\text{L}}$ attains the best Fidelity values consistently, with top results on Virus (Univar, $0.6859$), Scary (Token, $0.7617$), Daily (Univar, $0.7325$), Hipster (Multivar, $.7041$), and Trans (Multivar, $0.7645$). While Diversity values for both multi-token methods are slightly lower than those of $\text{DB}^{\text{L}}$, they remain competitive, particularly using the Token and Multivar generation methods.

No single generation method consistently outperforms others across all datasets. However, Token frequently provides higher Diversity, while Univar and Multivar strategies often achieve stronger Fidelity within individual datasets. 

Moreover, a detailed analysis of how multi-token training and the use of the regularization set affect these results can be found in the ablation study presented in Appendix~\ref{app1}.

In addition to these results, Table~\ref{tab:mean_runtime} reports the mean runtimes of the different training methods. The comparison highlights the effectiveness of LoRA in significantly reducing training time, as both $\text{DB}^{\text{L}}$ and $\text{DB}_{\text{MT}}^{\text{L}}$ complete in roughly two hours on average, compared to over a day for full DreamBooth training. In contrast, $\text{TI}$ exhibits the lowest runtime, since it avoids retraining the entire model and instead optimizes only the textual embeddings.

\begin{table}[h]
\centering
\caption{Mean runtime (minutes) comparison of the different training methods}
\resizebox{0.8\linewidth}{!}{
\begin{tabular}{l c c c c}
\toprule
 & $\text{TI}$ & $\text{DB}$ & $\text{DB}^{\text{L}}$ & $\text{DB}_{\text{MT}}^{\text{L}}$ \\
\midrule
Runtime (minutes) & $73.32 \pm 7.02$ & $2201.15 \pm 85.10$ & $135.56 \pm 9.42$ & $134.20 \pm 11.05$ \\
\bottomrule
\end{tabular}
}
\label{tab:mean_runtime}
\end{table}

\subsection{Approach 3. Human evaluation}\label{subsec-humanEval}

\begin{figure*}[htp]
    \centering
    \resizebox{\textwidth}{!}{%
        \includegraphics[]{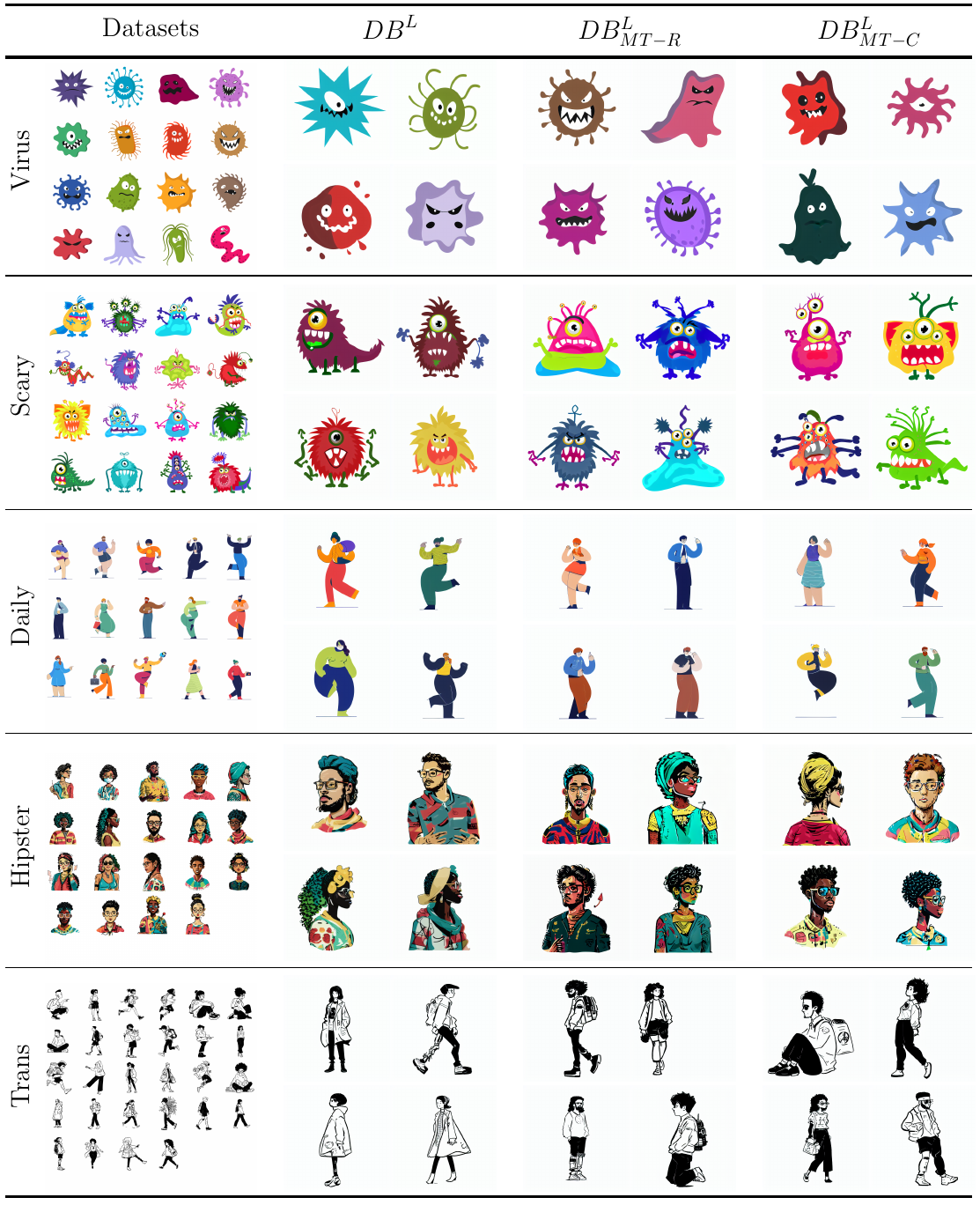}
    }
    \caption{Representative samples of generated characters shown to participants, organized by dataset and training method. Original datasets are included for reference.}
    \label{fig:genExamples}
\end{figure*}

The human evaluation followed the process outlined in Section \ref{subsubsec-humanEval}. $\text{DB}$ was excluded from the human evaluation due to its high invalidity rate, and $\text{TI}$ was also excluded given its combination of low Fidelity and high invalidity. To keep the comparisons in the human evaluation manageable, we selected a single generation strategy. Among the available methods, Multivar was chosen as it achieved the lowest proportion of invalid images ($6.16\%$) compared to Univar ($6.21\%$) and Token ($7.88\%$), while maintaining Fidelity values comparable to the others ($0.6955$ vs. $0.7100$ for Token and $0.7003$ for Univar) and offering competitive diversity ($0.2715$ vs. $0.2877$ for Token and $0.2691$ for Univar).
Figure~\ref{fig:genExamples} provides representative example of the images evaluated by participants. For this evaluation, only valid images from the selected methods and sampling strategy were included, ensuring that participants did not assess defective or otherwise invalid generations.

The results of the human evaluation are presented separately for the two participant groups: general participants and professional artists.

\vspace{6pt}\noindent\textit{General participants} performed pairwise comparisons, summarized in Table~\ref{tab:exp2NonExpert}. $\text{DB}_{\text{MT-C}}^{\text{L}}$ consistently ranks highest across all datasets, followed by $\text{DB}_{\text{MT-R}}^{\text{L}}$, and finally by $\text{DB}^{\text{L}}$. Notably, for the Trans dataset, general participants ranked images generated by both multi-token methods comparably or even preferable to images from the original dataset.

\vspace{6pt}\noindent\textit{Professional artists} numerically rated the generated images, as shown in Table~\ref{tab:exp3Artist}. Images directly from each original dataset (labeled as \textit{Dataset}) served as controls and received the highest global rating ($4.35$ out of $5$). Among evaluated methods, $\text{DB}_{\text{MT-C}}^{\text{L}}$ achieves the highest scores across all datasets (global average: $3.87$), followed by $\text{DB}_{\text{MT-R}}^{\text{L}}$ ($3.18$) and the baseline method $\text{DB}^{\text{L}}$ ($2.58$). Interestingly, on the \textit{Trans} dataset, $\text{DB}_{\text{MT-C}}^{\text{L}}$ surpasses the original dataset’s rating, consistent with the preference observed among general participants and further confirming the robustness of their evaluations.

\begin{table}[ht!]
\centering
\caption{Rankings obtained by general participants for different training methods. Each row represents a training method, with individual rankings provided for each dataset. A final Global ranking is included to summarize overall performance. Each value displays the ranking position along with the corresponding score for each method.}
\resizebox{\linewidth}{!}{
\setlength{\tabcolsep}{8pt}
\begin{tabular}{lrrrrrr}
\toprule
Method & Virus & Scary & Daily & Hipster & Trans & Global \\
\midrule
Dataset & \goldtrophy $ \ 1 \ (1082)$  & \goldtrophy $ \ 1 \ (1104)$ & \goldtrophy $ \ 1 \ (1086)$  & \goldtrophy $ \ 1 \ (1113)$   & \bronzetrophy $ \ 3 \ (1044)$   & \goldtrophy $ \ 1 \ (1083)$  \\[0.15em]
$\text{DB}_{\text{MT-C}}^{\text{L}}$& \silvertrophy $ \ 2 \ (1079)$   & \silvertrophy  $ \ 2 \ (1065)$  & \silvertrophy $ \ 2 \ (1014)$  & \silvertrophy $ \ 2 \ (1054)$   & \goldtrophy $ \ 1 \ (1053)$  & \silvertrophy $ \ 2 \ (1053)$    \\[0.15em]
$\text{DB}_{\text{MT-R}}^{\text{L}}$& \bronzetrophy $ \ 3 \ (988)$   & \bronzetrophy $ \ 3 \ (948)$   & \bronzetrophy $ \ 3 \ (969)$   & \bronzetrophy $ \ 3 \ (1009)$   & \silvertrophy $ \ 2 \ (1052)$  & \bronzetrophy $ \ 3 \ (994)$   \\[0.15em]
$\text{DB}^{\text{L}}$& \fourthtrophy $ \ 4 \ (851)$  & \fourthtrophy $ \ 4 \ (883)$ & \fourthtrophy $ \ 4 \ (932)$   & \fourthtrophy $ \ 4 \ (823)$   & \fourthtrophy $ \ 4 \ (851)$ & \fourthtrophy $ \ 4 \ (871)$   \\[0.15em]
\bottomrule
\end{tabular}
}
\label{tab:exp2NonExpert}
\end{table}




\begin{table}[ht!]
\centering
\caption{Rankings obtained by professional artists for different training methods. Each row represents a training method, with individual rankings provided for each dataset. A final Global ranking is included to summarize overall performance. Each value displays the ranking position along with the corresponding score for each method.}
\resizebox{\linewidth}{!}{
\setlength{\tabcolsep}{8pt}
\begin{tabular}{lrrrrrr}
\toprule
Method & Virus & Scary & Daily & Hipster & Trans & Global \\
\midrule
Dataset & \goldtrophy $ \ 1 \ (4.29)$  &  \goldtrophy $ \ 1 \ (4.25)$  &  \goldtrophy $ \ 1 \ (4.71)$  &  \goldtrophy $ \ 1 \ (4.30)$  &  \silvertrophy $ \ 2 \ (4.25)$  &  \goldtrophy $ \ 1 \ (4.35)$ \\[0.15em]
$\text{DB}_{\text{MT-C}}^{\text{L}}$ & \silvertrophy $ \ 2 \ (3.89)$  &  \silvertrophy $ \ 2 \ (3.62)$  &  \silvertrophy $ \ 2 \ (3.56)$  &  \silvertrophy $ \ 2 \ (3.82)$  &  \goldtrophy $ \ 1 \ (4.45)$  &  \silvertrophy $ \ 2 \ (3.87)$ \\[0.15em]
$\text{DB}_{\text{MT-R}}^{\text{L}}$ & \bronzetrophy $ \ 3 \ (3.10)$  &  \bronzetrophy $ \ 3 \ (2.65)$  &  \bronzetrophy $ \ 3 \ (3.20)$  &  \bronzetrophy $ \ 3 \ (3.27)$  &  \bronzetrophy $ \ 3 \ (3.76)$  &  \bronzetrophy $ \ 3 \ (3.18)$ \\[0.15em]
$\text{DB}^{\text{L}}$ & \fourthtrophy $ \ 4 \ (2.77)$  &  \fourthtrophy $ \ 4 \ (2.40)$  &  \fourthtrophy $ \ 4 \ (2.89)$  &  \fourthtrophy $ \ 4 \ (2.29)$  &  \fourthtrophy $ \ 4 \ (2.53)$  &  \fourthtrophy $ \ 4 \ (2.58)$ \\[0.15em]
\bottomrule
\end{tabular}
}
\label{tab:exp3Artist}
\end{table}

\section{Discussion}\label{sec-discussion}

The results of our study reveal differences in effectiveness across the evaluated methods. Starting with image validity (Section~\ref{subsec-filter}), the baseline method DreamBooth ($\text{DB}$) is shown to be unsuitable for style-consistent character generation, with nearly all outputs deemed invalid due to \textit{training copies}, reflecting overfitting. Textual Inversion ($\text{TI}$) performs slightly better in terms of validity, but still produces a large proportion of invalid samples, predominantly categorized as \textit{defective}. This suggests that, while TI avoids the overfitting failure mode of DreamBooth, its frozen-backbone training struggles to adapt to the combined demands of character identity and style consistency. Adding LoRA to DreamBooth ($\text{DB}^{\text{L}}$) reduces the number of invalid images compared with $\text{DB}$, but introduces other issues, most notably, a rise in \textit{defective images}. These often resemble examples from the regularization dataset, suggesting that the regularization process may interfere with the model's ability to generalize, particularly in low-data scenarios. The presence of \textit{multiple-subject} images, especially in datasets with human characters, further highlights the baseline methods’ limitations.

In contrast, the proposed multi-token methods ($\text{DB}_{\text{MT-R}}^{\text{L}}$ and $\text{DB}_{\text{MT-C}}^{\text{L}}$) demonstrate improvements in generation reliability, producing significantly fewer invalid images across all datasets. By explicitly separating character-specific tokens from the shared style token, these approaches reduce overfitting and unintended outputs, reinforcing the idea that such token disentanglement contributes to more stable and scalable generation.

Quantitative results (Section~\ref{subsec-comparison}) further support these findings. While Textual Inversion ($\text{TI}$) generates a moderate number of valid samples, it consistently shows lower Fidelity scores than all other methods, confirming that its learned embeddings fail to capture the target style and character details reliably. In contrast, the multi-token strategies better capture and preserve the artistic style present in the reference datasets. Between the two, the clustering-based variant ($\text{DB}_{\text{MT-C}}^{\text{L}}$) consistently achieves higher Fidelity scores than the rare-token variant ($\text{DB}_{\text{MT-R}}^{\text{L}}$), suggesting that maximizing token diversity in the embedding space, by clustering, improves the model’s ability to distinguish between individual character traits and shared stylistic features.

Diversity scores reveal further contrasts among the methods. Textual Inversion ($\text{TI}$) achieves the highest Diversity under the Token generation strategy but shows lower Diversity in the Univar and Multivar settings. However, this variety comes at the cost of substantially lower Fidelity, indicating that many outputs deviate from the intended style rather than providing meaningful stylistic variation. The DreamBooth-based methods show a different balance: adding LoRA to DreamBooth ($\text{DB}^{\text{L}}$) yields the highest Diversity among these approaches, but this is partly driven by generations with lower Fidelity, which often drift away from the style observed in the training data (see Figure~\ref{fig:Examples}). In contrast, the proposed multi-token methods produce slightly lower Diversity scores overall but maintain strong stylistic coherence, even in their lower-scoring outputs. This suggests that their diversity stems from controlled variation within the learned style. That said, the results of human evaluation (Section \ref{subsec-humanEval}) suggest that this decrease in Diversity did not negatively affect perceived character variety from a human perspective, suggesting that multi-token methods achieve a practically meaningful balance between maintaining style and generating novel characters.

Among generation strategies, no single method consistently dominated in both metrics, although the Univar and Multivar approaches offer the advantage of enabling virtually unlimited character generation. 

The human evaluation results (Section\ref{subsec-humanEval}) corroborate the quantitative performance comparison. General participants consistently preferred characters generated by $\text{DB}_{\text{MT-C}}^{\text{L}}$, closely followed by $\text{DB}_{\text{MT-R}}^{\text{L}}$, significantly ahead of $\text{DB}^{\text{L}}$. Moreover, for the Trans dataset, general participants rated characters from both multi-token methods even preferable to images from the original dataset, which highlights the perceived quality and stylistic coherence achieved by these approaches.

Professional artists’ evaluations further support these findings. As expected, the original reference images received the highest overall scores. Close behind, $\text{DB}_{\text{MT-C}}^{\text{L}}$ consistently achieved high ratings across all datasets, coming close to the quality of the reference images. Notably, in the Trans dataset, professional artists rated characters generated by $\text{DB}_{\text{MT-C}}^{\text{L}}$ even higher than the original images, reflecting the same preference observed among general participants. These results highlight the method’s strong stylistic integration and its practical value for artistic applications.

For a visual reference,  examples of high-Fidelity generations (Figure~\ref{fig:Examples}), along with the representative outputs (Figure~\ref{fig:genExamples}), may help to illustrate the stylistic consistency and character novelty achieved by the proposed method.

Despite the promising results and advantages of our proposed methods, several limitations remain. First, our approach is built on Stable Diffusion v1-5, meaning its effectiveness may be influenced by the underlying biases and constraints of this specific model. Second, the quantitative metrics (Fidelity and Diversity) are informative but imperfect, as they may not fully capture subtle stylistic details or clearly distinguish between meaningful character novelty and undesirable deviations. 

\section{Conclusion}\label{sec-conclusion}

In this work, we addressed the challenge of generating an unlimited number of novel characters that maintain stylistic consistency based on limited reference images. To overcome limitations of existing methods, we introduced a multi-token adaptation of DreamBooth combined with LoRA-based fine-tuning, explicitly distinguishing character-specific traits from shared stylistic features. 

Quantitative evaluations confirm the advantage of our clustering-based token selection strategy in achieving high stylistic fidelity. Human evaluations, both from general participants and professional artists, further validate these results. Our ablation study (\ref{app1}) further clarifies that multi-token training is the primary driver of Fidelity gains, while removing the regularization set eliminates major sources of invalidity, making their combination the most reliable and effective configuration.

Future research will explore the application of our proposed techniques with newer models such as Flux or Stable Diffusion 3.5. Additionally, we aim to develop refined metrics capable of capturing subtle stylistic differences, particularly when image quality is already high. Broader human evaluations will further complement these efforts, enhancing the reliability and practical relevance of character generation methods.

\appendix
\section{Ablation: Effect of the regularization set and Multi-Token training in $DB^L_{MT-C}$} \label{app1}

We ablate $\text{DB}^{\text{L}}_{\text{MT-C}}$ to isolate the effect of (i) removing the class-specific regularization set and (ii) enabling multi-token training. Table~\ref{tab:apendix1Ablation} reports results for four training variants: 
\begin{itemize}
    \item single-token with regularization ($\text{DB}^{\text{L}}$),
    \item multi-token with regularization ($\text{DB}^{\text{L}}_{\text{MT-C+Reg}}$),
    \item single-token model without regularization ($\text{DB}^{\text{L}}_{\text{noReg}}$), and 
    \item multi-token model without regularization ($\text{DB}^{\text{L}}_{\text{MT-C}}$).
\end{itemize}
As in the main text, we evaluate three generation strategies (Token/\hspace{0pt}Univar/\hspace{0pt}Multivar) on Virus, Scary, Daily, Hipster, and Trans, reporting invalid-image rate, Fidelity, and Diversity. All evaluations follow the methodology and experimental framework described in Section~\ref{subsec-baselines}, enabling a consistent comparison of how multi-token training and the regularization set influence each metric across datasets and generation strategies.

\begin{sidewaystable}
\centering
\caption{Ablation over training methods: single vs.\ multi-token and with vs.\ without a regularization set. For each training method combined with each generation strategy (Token\slash Univar\slash Multivar), we report Invalid (↓), Fidelity (↑), and Diversity (↑) on the five datasets. Best per-dataset/metric values are \textbf{bold}.}
\adjustbox{width=\textwidth, center}{
\begin{tabular}{llrrrrrrrrrrrrrrr}
\toprule
 &  & \multicolumn{3}{l}{Virus} & \multicolumn{3}{l}{Scary} & \multicolumn{3}{l}{Daily} & \multicolumn{3}{l}{Hipster} & \multicolumn{3}{l}{Trans} \\
 \cmidrule(r){3-5} \cmidrule(r){6-8} \cmidrule(r){9-11} \cmidrule(r){12-14} \cmidrule(r){15-17}
 Training & Generation  & Invalid ↓ & Fidelity ↑ & Diversity ↑ & Invalid ↓ & Fidelity ↑ & Diversity ↑ & Invalid ↓ & Fidelity ↑ & Diversity ↑ & Invalid ↓ & Fidelity ↑ & Diversity ↑ & Invalid ↓ & Fidelity ↑ & Diversity ↑ \\
\midrule
\multirow[t]{3}{*}{$DB^{{L}}$} & Token & $ 13.10\%$ & $ .6184 {\scriptscriptstyle \pm .0105}$ & $ \pmb{.3480 {\scriptscriptstyle \pm .0075}}$ & $ 3.00\%$ & $ .7241 {\scriptscriptstyle \pm .0066}$ & $ \pmb{.2945 {\scriptscriptstyle \pm .0061}}$ & $ 2.75\%$ & $ .7128 {\scriptscriptstyle \pm .0035}$ & $ .2741 {\scriptscriptstyle \pm .0041}$ & $ 55.75\%$ & $ .7038 {\scriptscriptstyle \pm .0061}$ & $ .3077 {\scriptscriptstyle \pm .0022}$ & $ 10.25\%$ & $ .7372 {\scriptscriptstyle \pm .0017}$ & $.3166 {\scriptscriptstyle \pm .0061}$ \\
 & Univar & $ 7.30\%$ & $ .6001 {\scriptscriptstyle \pm .0264}$ & $ .3422 {\scriptscriptstyle \pm .0122}$ & $ 0.80\%$ & $ .6894 {\scriptscriptstyle \pm .0195}$ & $ .2882 {\scriptscriptstyle \pm .0072}$ & $ 1.35\%$ & $ .7019 {\scriptscriptstyle \pm .0020}$ & $ .2736 {\scriptscriptstyle \pm .0068}$ & $ 64.50\%$ & $ .6969 {\scriptscriptstyle \pm .0088}$ & $ .2993 {\scriptscriptstyle \pm .0076}$ & $ 4.45\%$ & $ .7349 {\scriptscriptstyle \pm .0050}$ & $ .3056 {\scriptscriptstyle \pm .0016}$ \\
 & Multivar & $ 5.70\%$ & $ .6144 {\scriptscriptstyle \pm .0146}$ & $ .3437 {\scriptscriptstyle \pm .0095}$ & $ 1.75\%$ & $ .7097 {\scriptscriptstyle \pm .0129}$ & $ .2896 {\scriptscriptstyle \pm .0050}$ & $ 1.25\%$ & $ .6996 {\scriptscriptstyle \pm .0044}$ & $ .2730 {\scriptscriptstyle \pm .0078}$ & $ 60.10\%$ & $ .7035 {\scriptscriptstyle \pm .0027}$ & $ .2953 {\scriptscriptstyle \pm .0013}$ & $ 5.80\%$ & $ .7448 {\scriptscriptstyle \pm .0020}$ & $ .3041 {\scriptscriptstyle \pm .0016}$ \\
\midrule
\multirow[t]{3}{*}{$DB_{{MT-C+Reg}}^{{L}}$} & Token & $ 1.45\%$ & $ .6054 {\scriptscriptstyle \pm .0122}$ & $.3368 {\scriptscriptstyle \pm .0079}$ & $ 2.25\%$ & $ .7304 {\scriptscriptstyle \pm .0073}$ & $.2861 {\scriptscriptstyle \pm .0043}$ & $ 3.45\%$ & $ .7030 {\scriptscriptstyle \pm .0035}$ & $\pmb{.2757 {\scriptscriptstyle \pm .0030}}$ & $ 8.80\%$ & $ .7001 {\scriptscriptstyle \pm .0037}$ & $ .3104 {\scriptscriptstyle \pm .0022}$ & $ 7.65\%$ & $ .7278 {\scriptscriptstyle \pm .0037}$ & $\pmb{.3333 {\scriptscriptstyle \pm .0021}}$ \\
 & Univar & $ 1.30\%$ & $ .6223 {\scriptscriptstyle \pm .0097}$ & $ .3320 {\scriptscriptstyle \pm .0084}$ & $ 0.65\%$ & $ .7378 {\scriptscriptstyle \pm .0077}$ & $ .2723 {\scriptscriptstyle \pm .0055}$ & $ 3.40\%$ & $ .7063 {\scriptscriptstyle \pm .0023}$ & $ .2674 {\scriptscriptstyle \pm .0025}$ & $ 7.80\%$ & $ .6979 {\scriptscriptstyle \pm .0044}$ & $ .3065 {\scriptscriptstyle \pm .0025}$ & $ 8.85\%$ & $ .7260 {\scriptscriptstyle \pm .0049}$ & $ .3307 {\scriptscriptstyle \pm .0025}$ \\
 & Multivar & $ 1.90\%$ & $ .6178 {\scriptscriptstyle \pm .0097}$ & $ .3353 {\scriptscriptstyle \pm .0077}$ & $ 3.25\%$ & $ .7385 {\scriptscriptstyle \pm .0070}$ & $ .2771 {\scriptscriptstyle \pm .0030}$ & $ 5.80\%$ & $ .7034 {\scriptscriptstyle \pm .0034}$ & $ .2735 {\scriptscriptstyle \pm .0028}$ & $ 14.40\%$ & $ .6934 {\scriptscriptstyle \pm .0012}$ & $\pmb{.3132 {\scriptscriptstyle \pm .0023}}$ & $ 14.05\%$ & $ .7282 {\scriptscriptstyle \pm .0045}$ & $ .3329 {\scriptscriptstyle \pm .0019}$ \\
\midrule
\multirow[t]{3}{*}{$DB_{noReg}^{L}$} & Token & $\pmb{0.70\%}$ & $ .6642 {\scriptscriptstyle \pm .0038}$ & $ .3082 {\scriptscriptstyle \pm .0015}$ & $ 0.25\%$ & $ .7531 {\scriptscriptstyle \pm .0009}$ & $ .2671 {\scriptscriptstyle \pm .0022}$ & $ 0.45\%$ & $ .6982 {\scriptscriptstyle \pm .0030}$ & $ .2594 {\scriptscriptstyle \pm .0011}$ & $\pmb{0.70\%}$ & $ .6978 {\scriptscriptstyle \pm .0018}$ & $ .2805 {\scriptscriptstyle \pm .0018}$ & $ 0.75\%$ & $ .7476 {\scriptscriptstyle \pm .0020}$ & $ .3033 {\scriptscriptstyle \pm .0013}$ \\
 & Univar & $\pmb{0.70\%}$ & $ .6709 {\scriptscriptstyle \pm .0025}$ & $ .3053 {\scriptscriptstyle \pm .0026}$ & $ 0.40\%$ & $ .7521 {\scriptscriptstyle \pm .0019}$ & $ .2637 {\scriptscriptstyle \pm .0013}$ & $ 1.20\%$ & $ .7028 {\scriptscriptstyle \pm .0019}$ & $ .2541 {\scriptscriptstyle \pm .0021}$ & $ 0.75\%$ & $ .7009 {\scriptscriptstyle \pm .0008}$ & $ .2742 {\scriptscriptstyle \pm .0019}$ & $ 0.80\%$ & $ .7498 {\scriptscriptstyle \pm .0019}$ & $ .2964 {\scriptscriptstyle \pm .0011}$ \\
 & Multivar & $ 0.80\%$ & $ .6659 {\scriptscriptstyle \pm .0023}$ & $ .3075 {\scriptscriptstyle \pm .0016}$ & $ 0.20\%$ & $ .7524 {\scriptscriptstyle \pm .0011}$ & $ .2657 {\scriptscriptstyle \pm .0009}$ & $ 0.90\%$ & $ .7018 {\scriptscriptstyle \pm .0025}$ & $ .2590 {\scriptscriptstyle \pm .0013}$ & $ 0.85\%$ & $ .7010 {\scriptscriptstyle \pm .0005}$ & $ .2739 {\scriptscriptstyle \pm .0025}$ & $ 0.55\%$ & $ .7465 {\scriptscriptstyle \pm .0015}$ & $ .3018 {\scriptscriptstyle \pm .0017}$ \\
\midrule
\multirow[t]{3}{*}{$DB_{{MT-C}}^{{L}}$} & Token & $ 1.60\%$ & $ .6721 {\scriptscriptstyle \pm .0025}$ & $ .2846 {\scriptscriptstyle \pm .0013}$ & $ 0.20\%$ & $\pmb{.7617 {\scriptscriptstyle \pm .0014}}$ & $ .2473 {\scriptscriptstyle \pm .0010}$ & $ 1.35\%$ & $ .7149 {\scriptscriptstyle \pm .0025}$ & $ .2583 {\scriptscriptstyle \pm .0018}$ & $ 8.75\%$ & $ .7032 {\scriptscriptstyle \pm .0057}$ & $ .2870 {\scriptscriptstyle \pm .0015}$ & $ 2.55\%$ & $ .7541 {\scriptscriptstyle \pm .0056}$ & $ .3144 {\scriptscriptstyle \pm .0015}$ \\
 & Univar & $ 1.65\%$ & $\pmb{.6859 {\scriptscriptstyle \pm .0028}}$ & $ .2274 {\scriptscriptstyle \pm .0302}$ & $\pmb{0.00\%}$ & $ .7600 {\scriptscriptstyle \pm .0020}$ & $ .2214 {\scriptscriptstyle \pm .0146}$ & $ 5.35\%$ & $\pmb{.7331 {\scriptscriptstyle \pm .0031}}$ & $ .2528 {\scriptscriptstyle \pm .0030}$ & $ 1.00\%$ & $ .6944 {\scriptscriptstyle \pm .0102}$ & $ .2507 {\scriptscriptstyle \pm .0150}$ & $ 0.40\%$ & $ .7464 {\scriptscriptstyle \pm .0055}$ & $ .2945 {\scriptscriptstyle \pm .0077}$ \\
 & Multivar & $ 2.95\%$ & $ .6201 {\scriptscriptstyle \pm .0316}$ & $ .2737 {\scriptscriptstyle \pm .0065}$ & $ 0.05\%$ & $ .7596 {\scriptscriptstyle \pm .0057}$ & $ .2276 {\scriptscriptstyle \pm .0122}$ & $\pmb{0.05\%}$ & $ .7271 {\scriptscriptstyle \pm .0047}$ & $ .2419 {\scriptscriptstyle \pm .0096}$ & $\pmb{0.75\%}$ & $\pmb{.7041 {\scriptscriptstyle \pm .0030}}$ & $ .2662 {\scriptscriptstyle \pm .0115}$ & $ 0.45\%$ & $\pmb{.7645 {\scriptscriptstyle \pm .0061}}$ & $ .2860 {\scriptscriptstyle \pm .0141}$ \\
\bottomrule
\end{tabular}
}
\label{tab:apendix1Ablation}
\end{sidewaystable}

\paragraph{Multi-token with regularization ($\text{DB}^{\text{L}}$\ $\rightarrow$ $\text{DB}^{\text{L}}_{\text{MT-C+Reg}}$)}
Introducing multi-token training while keeping the regularization set
reduces invalid generations across datasets (e.g., Hipster–Univar $64.50\% \rightarrow 7.80\%$; Virus–Token $13.10\% \rightarrow 1.45\%$). Fidelity generally improves or remains comparable (e.g., Scary–Univar $0.6894 \rightarrow 0.7378$), while Diversity shows mixed but usually small shifts (slight decreases in several Token rows, occasional increases such as Hipster–Multivar $0.2953 \rightarrow 0.3132$). Thus, multi-token training primarily stabilizes generation and lifts Fidelity even in the presence of regularization.

\paragraph{Removing the regularization set  (single-token: $\text{DB}^{\text{L}}$\ $\rightarrow$ $\text{DB}^{\text{L}}_{\text{noReg}}$)}
Eliminating the regularization set
noticeably improves validity and Fidelity across datasets and generation strategies, with a general reduction in Diversity. For example, for Token generation, the invalid rate drops from $13.10\%$ to $0.70\%$ on \textit{Virus} and from $55.75\%$ to $0.70\%$ on Hipster, accompanied by consistent Fidelity gains (e.g., Virus $0.6184 \rightarrow 0.6642$, Scary $0.7241 \rightarrow 0.7531$). Similar trends hold for Univar and Multivar. These results suggest that the regularization set increases variability but introduces a mismatch that degrades alignment with the target distribution and increases the likelihood of invlid images. In short: removing the regularization set strongly boosts Fidelity and validity, at a modest cost in Diversity.

\paragraph{Combining both modifications ($\text{DB}^{\text{L}} \rightarrow \text{DB}^{\text{L}}_{\text{MT-C}}$)}
The variant used in the main results, multi-token with no regularization set, 
attains the strongest Fidelity across datasets, recovering the top per-dataset peaks observed earlier. Invalid rates remain low across strategies (typically below $3\%$, with only isolated raises). Diversity is generally lower than in the +Reg counterparts, which is consistent with the regularization set injecting variability, but remains competitive. Overall, multi-token training drives the Fidelity gains, while removing the regularization set eliminates a key source of invalidity and further improves style alignment.

\section*{Acknowledgments}

This research received support from an FPU grant (Formación de Profesorado Universitario) awarded by the Spanish Ministry of Science and Innovation (MCINN) to Rubén Pascual. 

This study was partially funded by the Spanish Ministry of Science and Innovation through the project PID2022-136627NB-I00 (MCIN/\hspace{0pt}AEI/\hspace{0pt}10.13039/\hspace{0pt}501100011033/\hspace{0pt}FEDER, UE).
\bibliographystyle{apalike}
\bibliography{invBibliography}
\end{document}